\documentclass[journal]{IEEEtran}

\usepackage[cmex10]{amsmath}
\usepackage{amssymb,amsfonts}
\usepackage{array}
\usepackage{multirow}
\usepackage{color}

\usepackage[colorlinks=true,citecolor=black,linkcolor=black,urlcolor=black]{hyperref}
\ifCLASSINFOpdf
  \usepackage[pdftex]{graphicx}
\else
  \usepackage[dvips]{graphicx}
\fi

\usepackage{subfigure}
\usepackage{float}
\usepackage{nccbbb}
\usepackage{galois}
\usepackage{url}
\usepackage{booktabs}
\usepackage{cite}

\usepackage[ruled, linesnumbered]{algorithm2e}

\newcommand{\etal}{\emph{et al. }}
\newcommand{\eg}{\emph{e.g. }}
\newcommand{\ie}{\emph{i.e. }}
\newcommand{\etc}{\emph{etc. }}

\begin{document}
%
\title{Semi-Supervised Sparse Representation Based Classification for Face Recognition with Insufficient Labeled Samples}
%
%
%

\author{Yuan Gao, Jiayi Ma, and Alan L. Yuille \emph{Fellow, IEEE}

\thanks{The authors would like to thank Weichao Qiu and Mingbo Zhao for giving feedbacks on the manuscript. This work was supported in part by the National Natural Science Foundation of China under Grant 61503288, in part by the China Postdoctoral Science Foundation under Grant 2016T90725, and in part by the NSF award CCF-1317376. \emph{(Corresponding author: Jiayi Ma.)}}
\thanks{Y. Gao is with the Electronic Information School, Wuhan University, Wuhan 430072, China, and also with the Tencent AI Laboratory, Shenzhen 518057, China. Email: Ethan.Y.Gao@gmail.com.}
\thanks{J. Ma is with the Electronic Information School, Wuhan University, Wuhan 430072, China. Email: jyma2010@gmail.com.}
\thanks{A. Yuille is with the Department of Statistics, University of California at Los Angeles, Los Angeles, CA 90095 USA, and also with the Department of Cognitive Science and the Department of Computer Science, Johns Hopkins University, Baltimore, MD 21218 USA. Email: yuille@stat.ucla.edu.}
}

%
%

\markboth{IEEE TRANSACTIONS ON IMAGE PROCESSING,~Vol.~XX, No.~XX, XXXX}%
{Gao \MakeLowercase{\textit{et al.}}: Semi-Supervised Gallery Dictionary Learning for Face Recognition}
%



\maketitle

\begin{abstract}
This paper addresses the problem of face recognition when there is only few, or even only a single, labeled examples of the face that we wish to recognize. Moreover, these examples are typically corrupted by nuisance variables, both linear (\ie additive nuisance variables such as bad lighting, wearing of glasses) and non-linear (\ie non-additive pixel-wise nuisance variables such as expression changes). The small number of labeled examples means that it is hard to remove these nuisance variables between the training and testing faces to obtain good recognition performance. To address the problem we propose a method called Semi-Supervised Sparse Representation based Classification (S$^3$RC). This is based on recent work on sparsity where faces are represented in terms of two dictionaries: a \emph{gallery dictionary} consisting of one or more examples of each person, and a \emph{variation dictionary} representing linear nuisance variables (\eg different lighting conditions, different glasses). The main idea is that (i) we use the variation dictionary to characterize the linear nuisance variables via the sparsity framework, then (ii) prototype face images are estimated as a gallery dictionary via a Gaussian Mixture Model (GMM), with mixed labeled and unlabeled samples in a semi-supervised manner, to deal with the non-linear nuisance variations between labeled and unlabeled samples. We have done experiments with insufficient labeled samples, even when there is only a single labeled sample per person. Our results on the AR, Multi-PIE, CAS-PEAL, and LFW databases demonstrate that the proposed method is able to deliver significantly improved performance over existing methods.
\end{abstract}


%
\IEEEpeerreviewmaketitle

\section{Introduction}
%
%
%
%
\IEEEPARstart{F}{ace} Recognition is one of the most fundamental problems in computer vision and pattern recognition. In the past decades, it has been extensively studied because of its wide range of applications, such as automatic access control system, e-passport, criminal recognition, to name just a few. Recently, the Sparse Representation based Classification (SRC) method, introduced by Wright \etal \cite{Wright09}, has received a lot of attention for face recognition \cite{Wagner12, Yang10, Yang11a, Ma15PR}. In SRC, a sparse coefficient vector was introduced in order to represent the test image by a small number of training images. Then the SRC model was formulated by jointly minimizing the reconstruction error and the $\ell^1$-norm on the sparse coefficient vector \cite{Wright09}. 
The main advantages of SRC have been pointed out in \cite{Wright10,Wright09}: i) it is simple to use without carefully crafted feature extraction, and ii) it is robust to occlusion and corruption.

One of the most challenging problems for practical face recognition application is the shortage of labeled samples \cite{tan2006face}. This is due to the high cost of labeling training samples by human effort, and because labeling multiple face instances may be impossible in some cases. For example, for terrorist recognition, there may be only one sample of the terrorist, \eg his/her ID photo. As a result, nuisance variables (or so called intra-class variance) can exist between the testing images and the limited amount of training images, \eg the ID photo of the terrorist (the training image) is a standard front-on face with neutral lighting, but the testing images captured from the crime scene can often include bad lighting conditions and/or various occlusions (\eg the terrorist may wear a hat or sunglasses). In addition, the training and testing images may also vary in expressions (\eg neutral and smile) or resolution. The SRC methods may fail in these cases because of the insufficiency of the labeled samples to model nuisance variables \cite{Shi11,Zhang11,Zhu10,Jiang14face,Jiang17srlsp}.

In order to address the insufficient labeled samples problem, Extended SRC (ESRC) \cite{Deng12} assumed that a testing image equals a prototype image plus some (linear) variations.
For example, a image with sunglasses is assumed to equal to the image without sunglasses plus the sunglasses. Therefore, ESRC introduced two dictionaries: (i) a gallery dictionary containing the prototype of each person (these are the persons to be recognized), and (ii) a variation dictionary which contains nuisance variations that can be shared by different persons (\eg different persons may wear the same sunglasses). Recent improvements on ESRC can give good results for this problem even when the subject only has a single labeled sample (namely the Single Labeled Sample Per Person problem, \ie SLSPP) \cite{Yang13,Zhuang13,Zhuang14,Wei15}.

However, various non-linear nuisance variables also exist in human face images, which makes prototype images hard to obtain. In other words, the nuisance variables often occur pixel-wise, which are not additive and cannot shared by different persons.
For example, we cannot simply add a specific variation to a neutral image (\ie the labeled training image) to get its smile images (\ie the testing images). Therefore, the limited number of training images may not yield a good prototype to represent the testing images, especially when non-linear variations exist between them. Attempts were to learn the gallery dictionary (\ie better prototype images) in Superposed SRC (SSRC) \cite{Deng13}. However, it requires multiple labeled samples per subject, and still used simple linear operations (\ie averaging the labeled faces \emph{w.r.t} each subject) to get the gallery dictionary. 

\begin{figure}[t]
\centering
\includegraphics[width=1\linewidth]{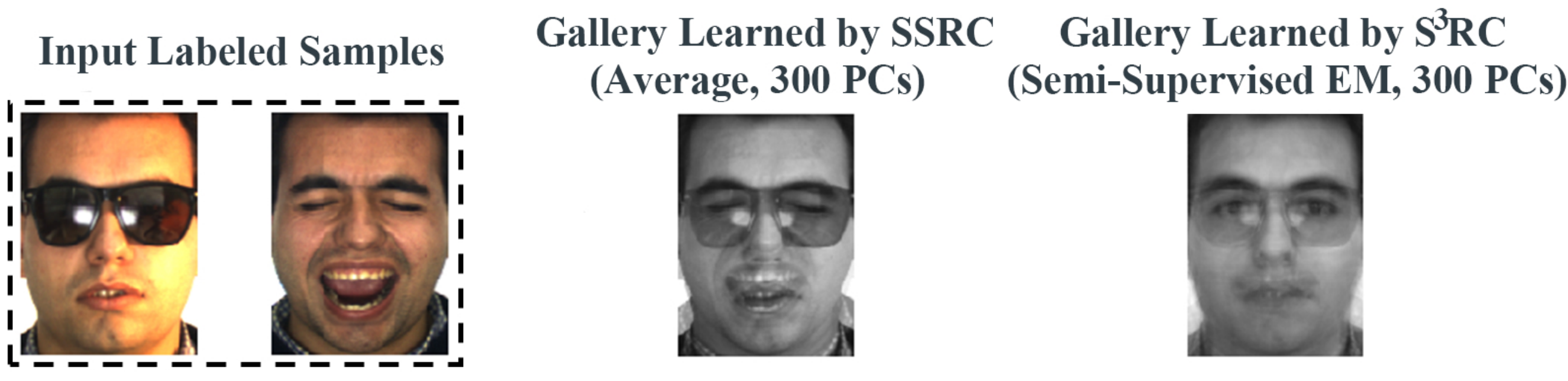}
\caption{Comparisons of the gallery dictionaries estimated by SSRC (\ie the mean of the labeled data) and our method (\ie one Gaussian centroid of GMM by semi-supervised EM initialized by the labeled data mean) using first 300 Principal Components (PCs, dimensional reduction by PCA). This illustrates that our method can estimate a better gallery dictionary with very few labeled images which contains both linear (\ie occlusion) and non-linear (\ie smiling) variations. The gallery from our method is learned by 5 semi-supervised EM iterations. 
}
\label{fig:centroid}
\end{figure}

In this paper, we propose a probabilistic framework called Semi-Supervised Sparse Representation based Classification (S$^3$RC) to deal with the insufficient labeled sample problem in face recognition, even when there is only one labeled sample per person. Both linear and non-linear variations between the training labeled and the testing samples are considered. We deal with the linear variations by a variation dictionary. After eliminated the linear variation (by simple subtraction), the non-linear variation is addressed by pursuing a better gallery dictionary (\ie better prototype images) via a Gaussian Mixture Model (GMM). 
Specifically, in our proposed S$^3$RC, the testing samples (without label information) are also exploited to learn a better model (\ie better prototype images) in a semi-supervised manner to eliminate the non-linear variation between the labeled and unlabeled samples. This is because the labeled samples are insufficient, and exploiting the unlabeled samples ensures that the learned gallery (\ie the better prototype) can well represent the testing samples and give better results. An illustrative example which compares the prototype image learned from our method and the existing SSRC is given in Fig. \ref{fig:centroid}. Clearly from Fig.\ref{fig:centroid}, we can see that, with insufficient labeled samples, a better gallery dictionary is learned by S$^3$RC that can well address the non-linear variations. Also Figs. \ref{fig:Illu_PIE} and \ref{fig:Illu_PIE2} in the later sections show that the learned gallery dictionary of our method can well represent the testing images for better recognition results.


In brief, since the linear variations can be shared by different persons (\eg different persons can wear the same sunglasses), therefore, we model the linear variations by a variation dictionary, where the variation dictionary is constructed by a large pre-labeled database which is independent of the training or testing. Then, we rectify the data to eliminate linear variations using the variation dictionary. After that, a GMM is applied to the rectified data, in order to learn a better gallery dictionary that can well represent the testing data which contains non-linear variation from the labeled training. Specifically, all the images from the same subject are treated as a Gaussian with its Gaussian mean as a better gallery. Then, the GMM is optimized to get the mean of each Gaussian using the semi-supervised Expectation-Maximization (EM) algorithm, initialized from the labeled data, and treating the unknown class assignment of the unlabeled data as the latent variable. Finally, the learned Gaussian means are used as the gallery dictionary for sparse representation based classification. The major contributions of our model are:
\begin{itemize}
\item Our model can deal with both linear and non-linear variations between the labeled training and unlabeled testing samples.
\item A novel gallery dictionary learning method is proposed which can exploit the unlabeled data to deal with the non-linear variations.
\item Existing variation dictionary learning methods are complementary to our method, \ie our method can be applied to other variation dictionary learning method to achieve improved performance.
\end{itemize}


The rest of the paper is organized as follows. We first summarize the notation and terminology in the next subsection. Section II describes background material and related work. SSRC and ESRC are described in Section III. In Section IV, starting with the insufficient training samples problem, we introduce the proposed S$^3$RC model, discuss the EM optimization, and then we extend S$^3$RC to the SLSPP problem. Extensive simulations have been conducted in Section V, where we show that by using our method as a classifier, further improved performance can be achieved using Deep Convolution Neural Network (DCNN) features. Section VI discusses the experimental results, and is followed by concluding remarks in Section VII.

\subsection{Summary of notation and terminology}
In this paper, capital bold and lowercase bold symbols are used to represent matrices and vectors, respectively. $\mathbf{1}_d \in {\mathbb{R}}^{d \times 1}$ denotes the unit column vector, and $\mathbf{I}$ is the identity matrix. $|| \cdot ||_1$, $|| \cdot ||_2$, $|| \cdot ||_F$ denote the $\ell^1$, $\ell^2$, and Frobenius norms, respectively. $\hat{a}$ is the estimation of parameter $a$.

In the following, we demonstrate the promising performance of our method on two problems with strictly limited labeled data: i) the insufficient uncontrolled gallery samples problem without generic training data, and ii) the SLSPP problem with generic training data. Here, uncontrolled samples are images containing nuisance variables such as different illumination, expression, occlusion, \etc We call these nuisance variables as intra-class variance in the rest of the paper. The generic training dataset is an independent dataset \emph{w.r.t} the training/testing dataset. It contains multiple samples per person to represent the intra-class variance. In the following, we use the insufficient training samples problem to refer to the former problem, and the SLSPP problem is short for the latter one. We do not distinguish the terms training/gallery/labeled samples, testing/unlabeled samples in the following. But note that the gallery samples and gallery dictionary are not identical. The latter means the learned dictionary for recognition.

The promising performance of our method is obtained by estimating the \emph{prototype} of each person as the gallery dictionary, and the prototype is estimated using both labeled and unlabeled data. Here, the \emph{prototype} means a learned image that represents the discriminative features of all the images from a specific subject. There is only one prototype for each subject. Typically, the prototype can be the neutral image of a specific subject without occlusion and obtained under uniform illumination. Our method learn the prototype by estimating the \emph{true centroid} for both labeled and unlabeled data of each person, thus we do not distinguish the \emph{prototype} and \emph{true centroid} in the following.

\section{Related Work}
The proposed method is a Sparse Representation based Classification (SRC) method. Many research works have been inspired by the original SRC method \cite{Wright09}. In order to learn a more discriminative dictionary, instead of using the training data itself, Yang \etal introduced the Fisher discrimination criterion to constrain the sparse code in the reconstructed error \cite{Yang11,Yang14}. Ma \etal learned another discriminative dictionary by imposing low-rank constraints on it \cite{Ma12}. Following these approaches, a model unifying \cite{Yang11} and \cite{Ma12} was proposed by Li \etal \cite{Li13,Li14}. Alternatively, Zhang \etal proposed a model to indirectly learn the discriminative dictionary by constraining the coefficient matrix to be low-rank \cite{Zhang13a}. Chi and Porikli incorporated SRC and Nearest Subspace Classifier (NSC) into a unified framework, and balanced them by a regularization parameter \cite{Chi12}. However, this category of methods need sufficient samples of each subject to construct an over-complete dictionary for modeling the variations of the uncontrolled samples \cite{Shi11,Zhang11,Zhu10}, and hence is not suitable for the insufficient training samples problem and the SLSPP problem.

Recently, ESRC was proposed to address the limitations of SRC when the number of samples per class is insufficient to obtain an over-complete dictionary, where a variation dictionary is introduced to represent the linear variation \cite{Deng12}. Motivated by ESRC, Yang \etal proposed the Sparse Variation Dictionary Learning (SVDL) model to learn the variation dictionary $\mathbf{V}$, more precisely \cite{Yang13}. In addition to modeling the variation dictionary by a linear illumination model, Zhuang \etal \cite{Zhuang14,Zhuang13} also integrated auto-alignment into their method. Gao \etal \cite{Gao14} extended the ESRC model by dividing the image samples into several patches for recognition. Wei and Wang proposed robust auxiliary dictionary learning to learn the intra-class variation \cite{Wei15}. The aforementioned methods did not learn a better gallery dictionary to deal with non-linear variation, therefore good prototype images (\ie the gallery dictionary) were hard to obtain. To address this issue, Deng \etal proposed SSRC to learn the prototype images as the gallery dictionary \cite{Deng13}. But this uses only simple linear operations to estimate the gallery dictionary, which requires sufficient labeled gallery samples and it is still difficult to model the non-linear variation.

There are semi-supervised learning (SSL) methods which use sparse/low-rank techniques. For example, Yan and Wang \cite{Yan09} used sparse representation to construct the weight of the pairwise relationship graph for SSL. He \etal \cite{He11} proposed a nonnegative sparse algorithm to derive the graph weights for graph-based SSL. Besides the sparsity property, Zhuang \etal \cite{Zhuang12,Zhuang15} also imposed low-rank constraints to estimate the weight matrix of the pairwise relationship graph for SSL. The main difference between them and our proposed method S$^3$RC is that the previous works used sparse/low-rank technologies to learn the weight matrix for graph-based SSL, which are essentially SSL methods. By contrast our method aims at learning a precise gallery dictionary in the ESRC framework, and the gallery dictionary learning was assisted by probability-based SSL (GMM), which is essentially a SRC method. Also note that as a general tool, GMM has been used for face recognition for a long time since Wang and Tang \cite{Wang12}. However, to the best of our knowledge, GMM has not been previously used for gallery dictionary learning in SRC based face recognitions.


\section{Semi-Supervised Sparse Representation based Classification with EM algorithm}
In this section, we present our proposed S$^3$RC method in detail. Firstly, we introduce the general SRC formulation with the gallery plus variation framework, in which the linear variation is directly modeled by the variation dictionary. Then, we prove that, after eliminating linear variations of each sample (which we call \emph{rectification}), the rectified data (both labeled and unlabeled) from one person can be modeled as a Gaussian to learn the non-linear variations. Following this, the whole rectified dataset including both labeled and unlabeled samples are formulated by a GMM. Next, initialized by the labeled data, the semi-supervised EM algorithm is used to learn the mean of each Gaussian as the \emph{prototype} images. Then, the learned gallery dictionary is used for face recognition by the gallery plus variation framework. After that, we describe the way to apply S$^3$RC to the SLSPP problem. Finally, the overall algorithm is summarized.

We use the \emph{gallery plus variation} framework to address both linear and non-linear variations. Specifically, the linear variation (such as illumination changes, different occlusions) is modeled by the variation dictionary. After eliminating the linear variation, we address the non-linear variation (\eg expression changes) between the labeled and unlabeled samples by estimating the centroid (prototype) of each Gaussian of the GMM. Note that GMM learn the class centroid (prototype) by semi-supervised clustering, \ie we only use the ground truth label as supervised information, the class assignment of the unlabeled data is treated as the latent variable in EM and updated iteratively during learning the class centroid (prototype).

\subsection{The gallery plus variation framework}
The SRC with gallery plus variation framework has been applied to the face recognition problem as follows. The observed images are considered as a combination of two different sub-signals, \ie a gallery dictionary $\mathbf{P}$ plus a variation dictionary $\mathbf{V}$ in the linear additive model \cite{Deng12}:
\begin{equation} \label{model}
\mathbf{y} = \mathbf{P} \mathbf{\alpha} + \mathbf{V} \mathbf{\beta} + \mathbf{e},
\end{equation}
where $\mathbf{\alpha}$ is a sparse vector that selects a limited number of bases from the gallery dictionary $\mathbf{P}$, and $\mathbf{\beta}$ is another sparse vector that selects a limited number of bases from the universal linear variation dictionary $\mathbf{V}$, and $\mathbf{e}$ is a small noise.

The sparse coefficients $\alpha, \beta$ can be estimated by solving the following $\ell^1$ minimization problem:
\begin{equation} \label{SSRC}
   \begin{bmatrix}
      \hat{\mathbf{\alpha}} \\
      \hat{\mathbf{\beta}}
   \end{bmatrix}
   = \mathop{\arg\min}_{\mathbf{\alpha}, \mathbf{\beta}}
   \begin{Vmatrix}
      \begin{bmatrix}
         \mathbf{P} & \mathbf{V}
      \end{bmatrix}
      \begin{bmatrix}
         \mathbf{\alpha} \\
         \mathbf{\beta}
      \end{bmatrix}
      - \mathbf{y}
   \end{Vmatrix}
   ^2_2 + \lambda
   \begin{Vmatrix}
      \begin{bmatrix}
         \mathbf{\alpha} \\
         \mathbf{\beta}
      \end{bmatrix}
   \end{Vmatrix}
   _1,
\end{equation}
where $\lambda$ is a regularization parameter. Finally, recognition can be conducted by calculating the reconstruction residuals for each class using $\hat{\alpha}$ (according to each class) and $\hat{\beta}$, \ie the test sample $\mathbf{y}$ is classified to the class with the smallest residual.

In this process, the linear additive variation (\eg illumination changes, different occlusions) of human faces can be directly modeled by the variation dictionary, given the fact that the linear additive variation can be shared by different subjects, \eg different persons may wear the same sunglasses. Let $\mathbf{A} = [\mathbf{A}_1, ..., \mathbf{A}_i, ..., \mathbf{A}_K] \in {\mathbb{R}}^{D \times n}$ denote a set of $n$ labeled images with multiple images per subject (class), where $\mathbf{A}_i \in {\mathbb{R}}^{D \times n_i}$ is the stacked $n_i$ sample vectors of subject $i$, and $D$ is the data/feature dimension, the (linear) variation dictionary can be constructed by:
\begin{equation}
    \mathbf{V} = [\mathbf{A}_{1}^{-}-\mathbf{a}_1^*\mathbf{1}_{n_1}^T, ..., \mathbf{A}_{K}^{-}-\mathbf{a}_K^*\mathbf{1}_{n_K}^T], \label{V1}
\end{equation}
\begin{equation}
    \mathbf{V} = [\mathbf{A}_{1}-\mathbf{c}_1 \mathbf{1}_{n_1}^T, ..., \mathbf{A}_{K}-\mathbf{c}_K \mathbf{1}_{n_K}^T], \label{V2}
\end{equation}
where $\mathbf{c}_i = \frac{1}{n_i} \mathbf{A}_i \mathbf{1}_{n_i} \in \mathbb{R}^{D \times 1}$ is the $i$-th class centroid of the labeled data. $\mathbf{a}_i^* \in \mathbb{R}^{D \times 1}$ is the \emph{prototype} of class $i$ that can \emph{best} represent the discriminative features of all the images from subject $i$, $\mathbf{A}_i^-$ is the complementary set of $\mathbf{a}_i^*$ according to $\mathbf{A}_i$.

The gallery dictionary $\mathbf{P}$ can then be set accordingly using one of the following equations:
\begin{equation}
    \mathbf{P} = \mathbf{A},  \label{P_ESRC}
\end{equation}
\begin{equation}
    \mathbf{P} = [\mathbf{c}_1, ...,  \mathbf{c}_K], \label{P}
\end{equation}




The aforementioned formulations of the gallery dictionary $\mathbf{P}$ and variation dictionary $\mathbf{V}$ works well when a large amount of labeled data is available. 
However, in practical applications such as recognizing a terrorist by his ID photo, the labeled/training data is often limited and the unlabeled/testing images are often taken under severely different conditions from the labeled/training data. Therefore, it is hard to obtain good prototype images to represent the unlabeled/testing images from the labeled/training data only. In order to address the non-linear variation between the labeled and unlabeled samples, in the following we learn a prototype $\mathbf{a}_i^*$ for each class by estimating the \emph{true centroid for both the labeled and unlabeled data of each subject}, and represent the gallery dictionary $\mathbf{P}$ using the learned prototype $\mathbf{a}_i^*$. (The importance of learning the gallery dictionary is shown in the previous Fig. \ref{fig:centroid} and Figs. \ref{fig:Illu_PIE} and \ref{fig:Illu_PIE2} in the later sections.)


\subsection{Construct the data from each class as a Gaussian after eliminating linear variations} \label{sec:PF}
We rectify the data to eliminate linear variations of each sample (\eg illumination changes, occlusions), so that the data from one person can be modeled as a Gaussian. This can be achieved by solving Eq.~\eqref{model} using Eq.~\eqref{P_ESRC} or \eqref{P} to represent the gallery dictionary and using Eq.~\eqref{V1} or \eqref{V2} to represent the variation dictionary:
\begin{equation}
\hat{\mathbf{y}} =  \mathbf{y} - \mathbf{V} \hat{\mathbf{\beta}} = \mathbf{P}\hat{\mathbf{\alpha}} + \mathbf{e}, \label{large_c}
\end{equation}
where $\hat{\mathbf{y}}$ is the rectified unlabeled image without linear variation, $\hat{\mathbf{\alpha}}$ and $\hat{\mathbf{\beta}}$ can be initialized by Eq. \eqref{SSRC}.

Then, the problem becomes to find the relationship between the rectified unlabeled data $\hat{\mathbf{y}}$ and its corresponding class centroid $\mathbf{a}^*$. Note that the sparse coefficient $\hat{\mathbf{\alpha}}$ is sparse and typically there is only one entry of $\mathbf{P}$ that represents each class. For an unlabeled sample, $\mathbf{y}$, $\mathbf{P}\hat{\mathbf{\alpha}}$ actually selects the most significant entry of $\mathbf{P}$, \ie, it selects the class centroid that is nearest to $\hat{\mathbf{y}}$.

However, the ``class centroid'' selected by $\mathbf{P}\hat{\mathbf{\alpha}}$ cannot be directly used as the initial class centroid for each Gaussian, because the biggest element of the sparse coefficient $\hat{\mathbf{\alpha}}$ typically does not take value 1. In other words, $\mathbf{P}\hat{\mathbf{\alpha}}$ can introduce scaling on the class centroid and additional (small) noise. More specifically, assume that the most significant entry of $\hat{\mathbf{\alpha}}$ is associated with class $i$, thus we have
\begin{align}
\mathbf{P}\hat{\mathbf{\alpha}} &= \mathbf{P}[\epsilon, \epsilon, ..., s~(i\text{-th entry}), ..., \epsilon]^T \nonumber \\
&= s \mathbf{a}_i^* + \mathbf{P} \hat{\mathbf{\alpha}}^{-}_{i} = s \mathbf{a}_i^* + \mathbf{e}',
\end{align}
where the sparse coefficient $\hat{\mathbf{\alpha}} = [\epsilon, \epsilon, ..., s~(i\text{-th entry}), ..., \epsilon]^T$ consisting of small values $\epsilon$ and a significant value $s$ in its $i$-th entry. $\mathbf{a}_i^*$ is the $i$-th column of $\mathbf{P}$, $\mathbf{e}' = \mathbf{P}\hat{\mathbf{\alpha}}^{-}_{i}$ is the summation of the ``noise'' class centroids selected by $\hat{\mathbf{\alpha}}^{-}_{i}$, in which $\hat{\mathbf{\alpha}}^{-}_{i}$  contains only the small values (\ie $\epsilon$'s) is the complementary set of $s$ according to $\hat{\mathbf{\alpha}}$. 

Recall that the gallery dictionary $\mathbf{P}$ has been normalized to have \emph{column} unit $\ell^2$-norm in Eq. \eqref{SSRC}, therefore, the scale parameter $s$ can be eliminated by normalizing $\mathbf{y} - \mathbf{V} \hat{\mathbf{\beta}}$ to have unit $\ell^2$-norm:
\begin{align}
\hat{\mathbf{y}}^{norm} &= norm(\mathbf{y} - \mathbf{V} \hat{\mathbf{\beta}}) = norm(s \mathbf{a}^* + \mathbf{e}' + \mathbf{e}) \nonumber \label{rectify1}\\
                        &\approx \mathbf{a}_i^* + \mathbf{e}^*,
\end{align}
where $\mathbf{e}^*$ is a small noise which is assumed to be a zero-mean Gaussian. Since there are insufficient samples from each subject, we assign the Gaussian noise of each class (subject) to be different from each other, \ie $\mathbf{e}_i^* = \mathcal{N}(0, \mathbf{\Sigma}_i)$, so as to estimate the gallery dictionary more precisely. Thus, the normalized $\hat{\mathbf{y}}$ obeys the following distribution:
\begin{equation}
\hat{\mathbf{y}}^{norm} \approx \mathbf{a}_i^* + \mathbf{e}^*_i \in \mathcal{N}(\mathbf{a}_i^*, \mathbf{\Sigma}_i). \label{Gaussian}
\end{equation}

\subsection{GMM Formulation}
After modeling the rectified data for each subject as a Gaussian, we construct a GMM for the whole data to estimate the \emph{true centroids} $\mathbf{a}^*$, to address the non-linear variation between the labeled/training and unlabeled/testing samples. Specifically, the unknown assignment for the unlabeled data is used as the latent variable. The detailed formulation is given in the following.

Let $\mathbf{D} = \{(\mathbf{y}_1, l_1) ..., (\mathbf{y}_n, l_n), \mathbf{y}_{n+1}, ..., \mathbf{y}_N\}$ denote a set of images of $K$ classes including both labeled and unlabeled samples, \ie $\{(\mathbf{y}_1, l_1) ..., (\mathbf{y}_n, l_n)\}$ are $n$ labeled samples with $\{ \mathbf{y}_i \in \mathbf{A}_{l_i}, i = 1, ..., n \}$; and $\{\mathbf{y}_{n+1}, ..., \mathbf{y}_N\}$ are $N-n$ unlabeled samples. Based on Eq. \eqref{Gaussian}, a GMM can be formulated to model the data and the EM algorithm can be used to more precisely estimate the true class centroids by clustering the normalized rectified images (that exclude the linear variations).

Firstly, the normalized rectified dataset $\hat{\mathbf{D}}^{norm} = \{(\hat{\mathbf{y}}^{norm}_1, l_1) ..., (\hat{\mathbf{y}}^{norm}_n, l_n), \hat{\mathbf{y}}^{norm}_{n+1}, ..., \hat{\mathbf{y}}^{norm}_N\}$ must be calculated in order to construct the GMM. The calculation of the normalized rectifications includes two parts: i) for the labeled data, the normalized rectifications are the roughly estimated class centroids; ii) for the unlabeled data, the normalized rectifications can be estimated by Eq. \eqref{rectify1}:
\begin{equation}
\hat{\mathbf{y}}^{norm}_i =
\left\{ \label{rectify}
   \begin{aligned}
      & \mathbf{c}_{l_i}, \quad \quad \quad \quad \quad \quad \quad  \text{if } i \in \{1, ..., n\}, \\
      & norm(\mathbf{y}_i - \mathbf{V} \hat{\mathbf{\beta}}_i), \quad  \text{if } i \in \{n+1, ..., N\},
   \end{aligned}
\right.
\end{equation}
where $\mathbf{c}_{l_i}$ is the mean of the labeled data of the $l_i$-th subject, \ie it is the roughly estimated centroid of class $l_i$, $\mathbf{V}$ is the variation dictionary, and $\hat{\mathbf{\beta}}_i$ is the sparse coefficient vector estimated by Eq. \eqref{SSRC}.

Following this, the GMM can be constructed as described in \cite{Zhu09}. Specifically, let $\pi_j$ denote the prior probability of class $j$, \ie $p(j) = \pi_j$, and $\theta$ be a set of unknown model parameter: $\theta = \{\mathbf{a}_j^*, \mathbf{\Sigma}_j, \mathbf{\pi}_j \text{, for } j = (1, ..., K) \}$. For the incomplete samples (the unlabeled data), an latent indicator $z_{i,j}$ is introduced to denote their label. That is, $z_{i,j} = 1$, if $\hat{\mathbf{y}}^{norm}_i \in \text{class } j$; otherwise $z_{i,j} = 0$.

Therefore, the objective function to optimize $\mathbf{\theta}$ can be obtained as:
\begin{equation}
\hat{\mathbf{\theta}} = \mathop{\arg\max}_{\mathbf{\theta}} \log p(\hat{\mathbf{D}}^{norm} | \theta), \quad \text{s.t. } \sum_j^K \pi_j = 1,
\end{equation}
where the log likelihood $\log p(\hat{\mathbf{D}}^{norm} | \theta)$ is:
\begin{align}
  &\log p(\hat{\mathbf{D}}^{norm} | \theta) = \log \prod_{i=1}^{N} p(\hat{\mathbf{y}}^{norm}_i | \theta)^{z_{i,j}} \nonumber \\
 =& \log \left(\prod_{i=1}^n p(\hat{\mathbf{y}}^{norm}_i, l_i | \theta) \prod_{i=n+1}^{N} p(\hat{\mathbf{y}}^{norm}_i | \theta)^{z_{i,j}} \right) \nonumber \\
 =& \sum_{i=1}^n \log \pi_{l_i} \mathcal{N}(\hat{\mathbf{y}}^{norm}_i | \mathbf{a}_{l_i}^*, \mathbf{\Sigma}_{l_i}) \quad + \label{likelihood} \nonumber\\
 & \sum_{i=n+1}^N \sum_{j=1}^K z_{i,j} \log\pi_j \mathcal{N}(\hat{\mathbf{y}}^{norm}_i | \mathbf{a}_i^*, \mathbf{\Sigma}_i).
\end{align}
In Eq. \eqref{likelihood} the label of sample $i$, \ie $l_i$, was used as a subscript to index the mean and variance, \ie $\mathbf{a}_{l_i}^*, \mathbf{\Sigma}_{l_i}$, of the cluster which sample $i$ belongs to.

\subsection{Estimation of the gallery dictionary by semi-supervised EM algorithm}
The EM algorithm is applied to estimate the unknown parameters $\theta$ by iteratively calculating the latent indicator (label) of the unlabeled data and maximizing the log likelihood $\log p(\hat{\mathbf{D}}^{norm} | \theta)$ \cite{ma2015robust,Gao14_nonrigid,Ma13PR,Ma16}.

For the EM iterations, $\mathbf{a}_j^*$, $\Sigma_j$ and $\pi_j$ for $j = 1,...,K$ are initialized by $\mathbf{p}_j$, $\mathbf{I}$, and $n_i / n$, respectively. Here, $n_i$ is the number of labeled samples in each class and $n$ is the total number of labeled samples.

\textbf{E-Step}: this aims at estimating the latent indicator, $z_{i,j}$, of the unlabeled data using the current estimate of $\theta$. For the labeled data, $z_{i,j}$ is known then it is set to its label, which is the main difference between semi-supervised EM and original EM. (This has already been applied in Eq. \eqref{likelihood}):
\begin{equation}
\hat{z}_{i,j} =
\left\{ \label{z_label}
   \begin{aligned}
       &1, \qquad \quad \text{if } i \in \{1, ..., n\} \text{ and } j = l_i, \\
       &0, \qquad \quad \text{if } i \in \{1, ..., n\} \text{ and } j \neq l_i. \\
   \end{aligned}
\right.
\end{equation}
Equation \eqref{z_label} ensures that the ``estimated labels'' of the labeled samples are fixed by their true labels. Thus, the labeled data plays a role of anchors, which encourage the EM, applied to the whole dataset, to converge to the true gallery.

For the unlabeled data, the algorithm uses the expectation of $z_{i,j}$ from the previously estimated model $\mathbf{\theta}^{old}$, to give a good estimation of $z_{i,j}$:
\begin{align}
\hat{z}_{i,j} &= \mathbb{E}[z_{i,j}] \nonumber \\
&= \frac{ \pi_j^{old} \frac{1}{|\Sigma_j^{old}|^{1/2}}  \exp{(-\frac{1}{2} ||\hat{\mathbf{y}}^{norm}_i - (\mathbf{a}_j^{*})^{old}||^2_{\Sigma_j^{old}} )} }{ \sum_{k=1}^K \pi_k^{old}  \frac{1}{|\Sigma_k^{old}|^{1/2}} \exp{(-\frac{1}{2} ||\hat{\mathbf{y}}^{norm}_i - (\mathbf{a}_k^{*})^{old}||^2_{\Sigma_k^{old}}) }}, \nonumber \label{z_unlabel} \\
&\text{if } i \in \{n+1, ..., N\}.
\end{align}

\textbf{M-Step}: $z_{i,j}$ in Eq. \eqref{likelihood} can be substituted by $\hat{z}_{i,j}$, so that we can optimize the model parameter $\mathbf{\theta}$. By using Maximum-Likelihood Estimation (MLE), the optimized model parameters can be obtained by:
\begin{align}
&\hat{N}_j = \sum_{i=1}^{N} \hat{z}_{i,j}, \qquad \hat{\pi}_j = \frac{\hat{N}_j }{N}. \label{pi} \\
&\hat{\mathbf{a}}_j^* = \frac{1}{\hat{N}_j}\sum_{i=1}^{N}\hat{z}_{i,j}\hat{\mathbf{y}}^{norm}_i, \label{mu} \\
&\hat{\Sigma}_j = \frac{1}{\hat{N}_j}\sum_{i=1}^{N}\hat{z}_{i,j}(\hat{\mathbf{y}}^{norm}_i - \hat{\mathbf{a}}_j^*)(\hat{\mathbf{y}}^{norm}_i - \hat{\mathbf{a}}_j^*)^T, \label{Sigma}
\end{align}

The E-Step and M-Step iterate until the log likelihood $\log p(\hat{\mathbf{D}}^{norm} | \theta)$ converges.

\textbf{Initialization}: The EM algorithm needs good initialization to avoid the local minima. To initialize the mean of each Gaussian $\mathbf{a}_i^*$, it is natural to use the roughly estimated class centroids, \ie the mean of labeled data, $\mathbf{c}_i$ (See Fig.\ref{fig:centroid} as an example). The variance for each class is initialized by the identity matrix, \ie $\Sigma_i = \mathbf{I}, i = 1,...,K$.

\begin{figure*}[t]
\centering
\includegraphics[width=0.7\linewidth]{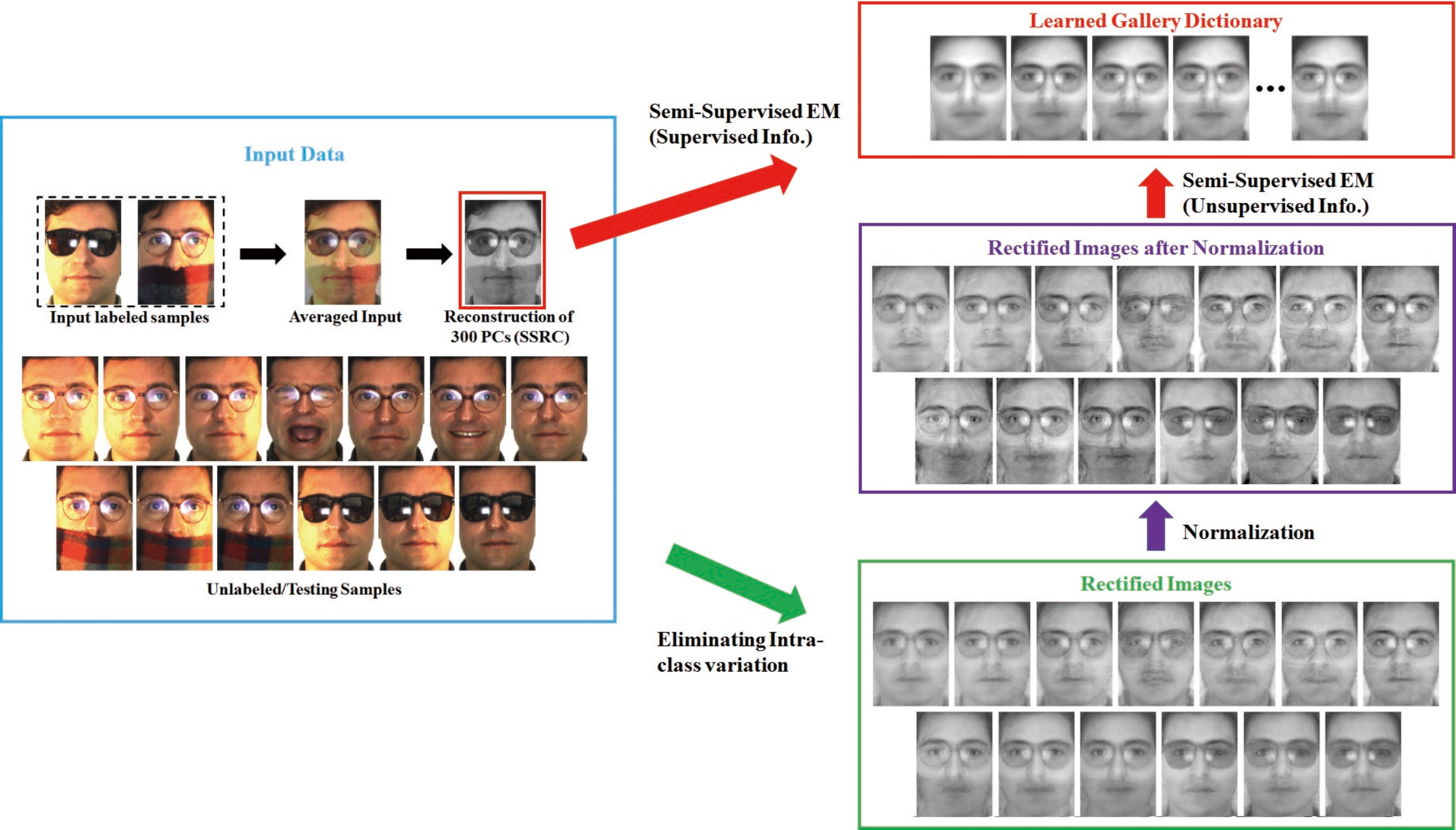}
\caption{The illustration of the procedures of the proposed semi-supervised gallery dictionary learning method.}
\label{fig:illu_S$^3$RC_All}
\end{figure*}

\subsection{Classification using the estimated gallery dictionary and SRC} \label{sec:classification}
The estimated $\hat{\mathbf{a}}_j^*$ is used as the gallery dictionary, $\mathbf{P}^* = [\hat{\mathbf{a}}_1^*, ..., \hat{\mathbf{a}}_K^*]$. Then $\mathbf{P}^*$ is used to substitute $\mathbf{P}$ in Eq. \eqref{SSRC}, to estimate the new sparse coefficients, $\mathbf{\alpha}^*$ and $\mathbf{\beta}^*$. Finally, the residuals for each class $k$ are computed for the final classification by:
\begin{equation}
r_k(\mathbf{y}) = \label{residual}
   \begin{Vmatrix}
      \mathbf{y} -
      \begin{bmatrix}
         \mathbf{P^*} & \mathbf{V}
      \end{bmatrix}
      \begin{bmatrix}
         \delta_k(\hat{\alpha}^*) \\
         \hat{\beta}^*
      \end{bmatrix}
   \end{Vmatrix}
_F^2,
\end{equation}
where $\delta_k(\hat{\alpha}^*)$ is a vector whose nonzero entries are the entries in $\hat{\alpha}^*$ that are associated with class $k$. Then the testing sample $\mathbf{y}$ is classified into the class with smallest $r_k(\mathbf{y})$, \ie Label$(\mathbf{y}) = \arg\min_k r_k(\mathbf{y})$.

\subsection{S$^3$RC model for SLSPP problem with generic dataset}
Recently, a lot of researchers have introduced an extra generic dataset for addressing the face recognition problem with SLSPP \cite{Deng12,Gao14,Kan13,Su10,Yang13}, of which the SRC methods \cite{Deng12,Gao14,Yang13} have achieved state-of-the-art results. Here, the generic dataset can be an independent dataset from the training and testing dataset $\mathbf{D}$. When a generic dataset is given in advance, our model can also be easily applied to the SLSPP problem.

In the SLSPP problem, the input data has $N$ samples, $\mathbf{D} = \{(\mathbf{y}_1, 1),  ..., (\mathbf{y}_K, K), \mathbf{y}_{K+1}, ..., \mathbf{y}_N\}$. From $\mathbf{D}$, the set of the labeled data, $\mathbf{T} = \{\mathbf{y}_1,  ..., \mathbf{y}_K\}$, is known as the gallery dataset, where there is only one sample for each subject. $\mathbf{G} = \{\mathbf{G}_1, ..., \mathbf{G}_{K^g} \} \in \mathbb{R}^{D \times N^g}$ denotes a labeled generic dataset with $N^g$ samples and $K^g$ subjects in total. Here, $D$ is the data dimension shared by gallery, generic and testing data, and $\mathbf{G}_i \in {\mathbb{R}}^{D \times n_i^g}$ is the stacked $n_i^g$ vectors of the samples from class $i$.

Due to the limited number of gallery samples, the initial class center is set to the only labeled sample of each class, and the corresponding variation dictionary can be constructed similar to Eq. \eqref{V2}:
\begin{align}
&\mathbf{P} = \mathbf{T}, \label{P_SLSPP_1}\\
&\mathbf{V} = [\mathbf{G}_{1}-\mathbf{c}_1^g \mathbf{1}_{n_1^g}^T, ..., \mathbf{G}_{K}-\mathbf{c}_1^g \mathbf{1}_{n_{K^g}^g}^T], \label{V_SLSPP_2}
\end{align}
where $\mathbf{c}_i^g$ is the average of the samples according to $i$-th subject in generic dataset, \ie $\mathbf{c}_i^g = \frac{1}{n_i^g} \mathbf{G}_i \mathbf{1}_{n_i^g}$.

The obtained initial gallery dictionary and variation dictionary can then be applied in our model, as discussed in Section \ref{sec:PF} -- Section \ref{sec:classification}.

\subsection{Summary of the algorithm}
The overall algorithm is summarized in Algorithm \ref{algorithm}. We also gave an illustrated procedures of the proposed method in Fig. \ref{fig:illu_S$^3$RC_All}. The inputs to the algorithm, besides the regularization parameter $\lambda$, are:
\begin{itemize}
\item For the insufficient training samples problem: a set of images including both labeled and unlabeled samples $\mathbf{D} = \{(\mathbf{y}_1, l_1)  ..., (\mathbf{y}_n, l_n), $    $ \mathbf{y}_{n+1}, ..., \mathbf{y}_N\}$, where $\mathbf{y}_i \in \mathbb{R}^D, i = 1, ..., N$ are image vectors, and $l_i, i = 1,...,n$ are the labels. We denote $n_i$ to be the number of labeled samples of each class.
\item For the SLSPP problem with generic dataset: the dataset including gallery and testing data $\mathbf{D} = \{(\mathbf{y}_1, 1)  ..., (\mathbf{y}_K, K), \mathbf{y}_{K+1}, ..., \mathbf{y}_N\}$, in which $\mathbf{T} = \{\mathbf{y}_1,  ..., \mathbf{y}_K \}$ is the gallery set with SLSPP and 1, ..., $K$ are the labels. A labeled generic dataset with $N^g$ samples from $K^g$ subjects, $\mathbf{G} = \{\mathbf{G}_1, ..., \mathbf{G}_{N^g}\}$.
\end{itemize}

\begin{algorithm}[!thp]
\SetNlSkip{0.5em}
\SetInd{0.5em}{1em}
\caption{Semi-Supervised Sparse Representation based Classification}
\label{algorithm}
Compute the prototype matrix, $\mathbf{P}$, by Eq. \eqref{P} (for the insufficient training samples problem), or by Eq. \eqref{P_SLSPP_1} (for the SLSPP problem). \\
Compute the universal linear variation matrix, $\mathbf{V}$, by Eq. \eqref{V2} (for the insufficient training samples problem), or by Eq. \eqref{V_SLSPP_2} (for the SLSPP problem). \\
\tcc{$\mathbf{V}$ can also be calculated by other variation dict. learning methods.} 
Apply dimensional reduction (\eg PCA) on the whole dataset as well as $\mathbf{P}$ and $\mathbf{V}$, and then normalize them to have column unit $\ell^2$ norm.\\
Solve the Sparse Representation problem to estimate $\hat{\alpha}$ and $\hat{\beta}$ for all the unlabeled $\mathbf{y}$ by Eq. \eqref{SSRC}. \\
Rectify the samples to eliminate linear variation and normalize them to have unit $\ell^2$-norm by Eq. \eqref{rectify}. \\
Initialize each Gaussian of the GMM by $\mathcal{N}(\mathbf{p}_i, \mathbf{I})$ for $i = 1, ... K$, where $\mathbf{p}_i$ is the $i$-th column of $\mathbf{P}$. \\
Initialize the prior of GMM, $\pi_i = n_i/n$  (for the insufficient training samples problem), or $\pi_i = 1/K$ (for SLSPP problem) for $i = 1, ... K$. \\
\Repeat{\rm Eq. \eqref{likelihood} converges}{
\textbf{E-Step}: Calculate $\hat{z}_{ij}$ by Eqs. \eqref{z_label} and \eqref{z_unlabel}. \\
\textbf{M-Step}: Optimize the model parameter $\theta = \{\mathbf{\mu}_j, \mathbf{\Sigma}_j, \mathbf{\pi}_j, \text{ for } j=1,...,K \}$ by Eqs. \eqref{pi}--\eqref{Sigma}.
}
Let $\mathbf{P}^* = [\hat{\mu}_1, ..., \hat{\mu}_K]$, estimate $\hat{\mathbf{\alpha}^*}, \hat{\mathbf{\beta}^*}$ by Eq. \eqref{SSRC}. \\
Compute the residual $r_k(\mathbf{y})$ by Eq. \eqref{residual}. \\
\textbf{Output:} Label$(\mathbf{y}) = \arg\min_k r_k(\mathbf{y})$.
\end{algorithm}

\begin{figure}[b]
\centering
\includegraphics[width=0.6\linewidth]{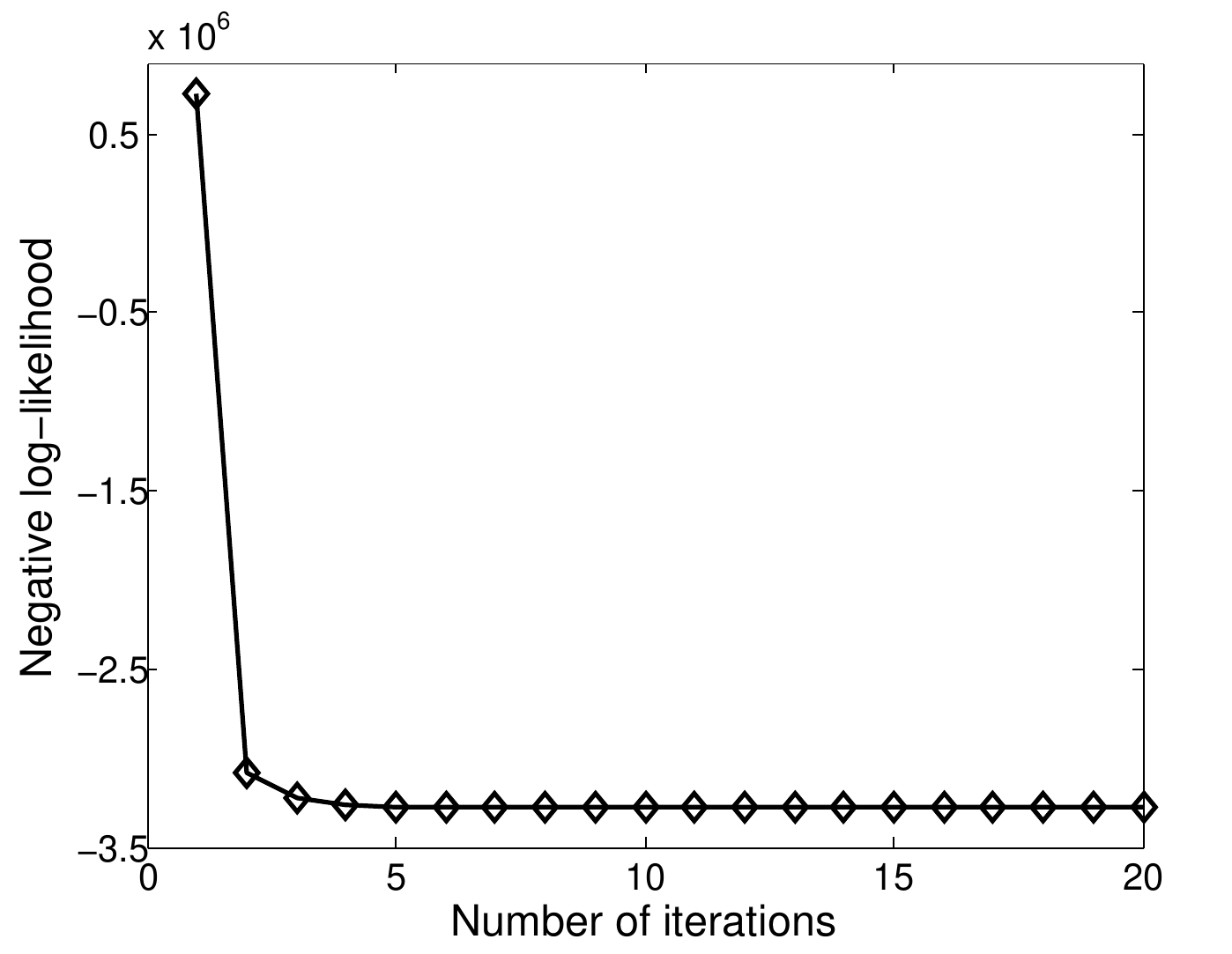}
\caption{The typical performance of the change in negative log-likelihood (from a randomly selected trial on the AR database), which convergence with less than 5 iterations.}
\label{fig:likelihood}
\end{figure}

To investigate the convergence of the semi-supervised EM, Fig. \ref{fig:likelihood} illustrates the typical performance of change in negative log-likelihood (from a randomly selected run on the AR database), which convergence with less than 5 iterations. A converged example is illustrated in Fig. \ref{fig:centroid} (\ie the rightmost subfigure, obtained by 5 iterations). From both Figs. \ref{fig:centroid} and \ref{fig:likelihood}, it can be observed that the algorithm converges quickly and the discriminative features of each specific subject can be learned in a few steps.

Note that our algorithm is related to the SRC methods which also used the \emph{gallery plus variation} framework, such as ESRC \cite{Deng12}, SSRC \cite{Deng13}, SVDL \cite{Yang13} and SILT \cite{Zhuang14, Zhuang13}. Among these method, only SSRC aims to learn the gallery dictionary, but it needs sufficient label/training data. Also it is not ensured that the learned prototype from SSRC can well represent the testing data due to the possible severe variation between the labeled/training and the testing samples. While ESRC, SVDL and SILT focus to learn a more representative variation dictionary. The variation dictionary learning in these methods are complementary to our proposed method. In other words, we can use them to replace Eqs. \eqref{V1}, \eqref{V2} or \eqref{V_SLSPP_2} for better performance, which will be verified in the experiments by using SVDL to construct our variation dictionary.

\section{Results}
In this section, our model is tested to verify the performance for both the insufficient training samples problem and the SLSPP problem. Firstly, the performance of the proposed method on the insufficient training samples problem using the AR database \cite{Martinez98} is shown. Then, we test our method on the SLSPP problem on both the Multi-PIE \cite{Gross09} and the CAS-PEAL \cite{Gao08} databases, using one facial image with neutral expression from each person as the only labeled gallery sample. Next, in order to further investigate the performance of the proposed semi-supervised gallery dictionary learning, we re-do the SLSPP experiments with one randomly selected image (\ie uncontrolled image) per person as the only gallery sample. The performance has been evaluated on the Multi-PIE \cite{Gross09}, and the more challenging LFW \cite{LFWTech} databases. After that, the influence of different amounts of labeled and/or unlabeled data is investigated. Finally, we have illustrated the performance of our method through a practical system with automatic face detection and automatic face alignment.

For all the experiments, we report the results from both \emph{transductive} and \emph{inductive} experimental settings. Specifically, in the transductive setting, the data is partitioned into two parts, \ie the labeled training and the unlabeled testing. We focus on the performance on the unlabeled/testing data, where we do not distinguish the unlabeled and the testing data. In inductive setting, we split the data into three parts, \ie the labeled training, the unlabeled training and the testing, where the model is learned by the labeled training and the unlabeled training data, the performance is evaluated on the testing data. In order to provide comparable results, the Homotopy method \cite{Asif10,Donoho08,Osborne00} was used to solve the $\ell^1$ minimization problem for all the methods involved.

\subsection{Performance on the insufficient training samples problem}
\begin{figure}[!htp]
\centering
\includegraphics[width=0.9\linewidth]{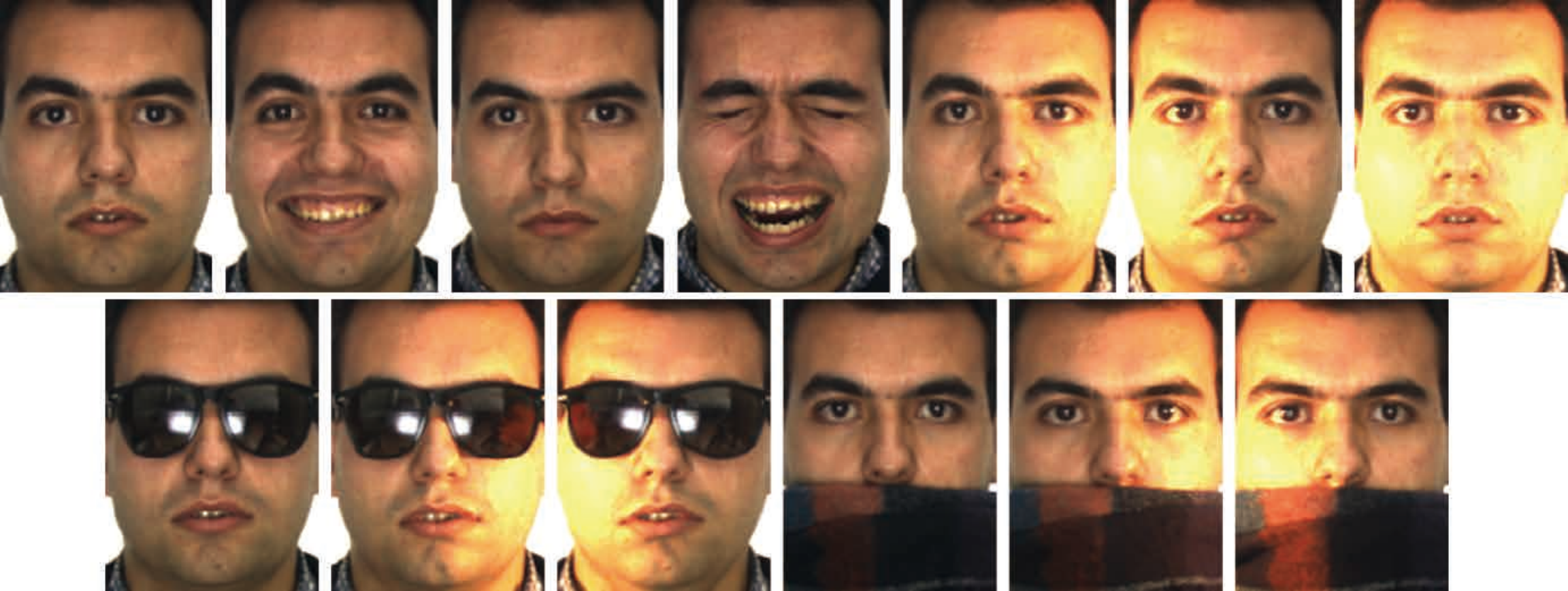}
\caption{The cropped image instances of one subject from the AR database (one complete session).}
\label{fig:ARinstance}
\end{figure}

In the AR database \cite{Martinez98}, there are 126 subjects with 26 images for each of them taken at two different sessions (dates), each session containing 13 images. The variations in the AR database include illumination change, expressions and facial occlusions. In our experiment, a cropped and normalized subset of the AR database that contains 100 subjects with 50 males and 50 females is used. The corresponding 2600 images are cropped to $165 \times 120$. This subset of AR used in our experiment has been selected and cropped by the data provider \cite{Martinez01}\footnote{The data can be downloaded at \url{http://cbcsl.ece.ohio-state.edu/protected-dir/AR_warp_zip.zip} upon authorization.}. Figure \ref{fig:ARinstance} illustrates one session (13 images) for a specific subject. 

Here we conduct four experiments to investigate the performance of our methods, and the results are reported from both the transductive and the inductive settings. No extra generic dataset is used in either of the experiments. The experimental settings are:
\begin{itemize}
	\item \emph{Transductive}: There are two transductive experiments. One is an extension of the experiment in \cite{Deng13,Shi11}, by using different amounts of labeled data. In the experiment, 2-13 uncontrolled images (of 26 images in total) of each subject are randomly chosen as the labeled data, and the remaining 24-13 images are used as unlabeled/query data for EM clustering and testing. The two sessions are not separated in this experiment. The results of this experiment are shown in Figure \ref{fig:AR_Results}a. Figure \ref{fig:AR_Results}b is a more challenging experiment, in which the labeled and unlabeled images are from different sessions. More explicitly, we first randomly choose a session, then randomly select 2-13 images of each subject from that session to be the labeled data. The remaining 13 images from the other sessions are used as unlabeled/query data.
	\item \emph{Inductive}: There are also two inductive experiments as shown in Figs. \ref{fig:AR_Results}c and \ref{fig:AR_Results}d, where the labeled training is selected by the same strategy used in the transductive settings. Specifically, the experiment in Fig. \ref{fig:AR_Results}c uses 2-13 (of 26 images in total) randomly selected uncontrolled images as labeled training samples. Thereafter, the remaining 24-13 images are randomly separated into two parts, \ie half as unlabeled training samples and the other half as testing samples. Similarly, Fig. \ref{fig:AR_Results}d shows another more challenging experiment, where the training (including both labeled and unlabeled) and the testing samples are chosen from different sessions to ensure the large differences between them. That is, for each person, a session is randomly selected at first. From that session, then, 2-12 randomly selected images are used as the labeled training samples, and the remaining 11-1 images are used as the unlabeled training samples. All the 13 images from the other session are used as testing samples.
\end{itemize}

State-of-the-art methods for the insufficient training samples problem are used for comparison, including sparse/dense representation methods SRC \cite{Wright09} and ProCRC \cite{Caiprobabilistic16}, low-rank models DLRD\_SR \cite{Ma12}, $D^2 L^2 R^2$ \cite{Li13,Li14}, gallery plus variation representation ESRC \cite{Deng12}, SSRC \cite{Deng13}, and RADL \cite{Wei15}. We follow the same settings of other parameters as in \cite{Deng13}, \ie first 300 PCs from PCA \cite{Bishop06} have been used and $\lambda$ is set to 0.005. The results are illustrated by the mean and standard deviation of 20 runs. Particularly, in order to show that the generalizability of the proposed gallery dictionary learning S$^3$RC, we also investigate the performance of our method in RADL framework, \ie we replace the gallery dictionary in RADL by the learned gallery from S$^3$RC and keep the remaining part of the RADL model unchanged. We denote it by S$^3$RC-RADL.

\begin{figure}[!htp]
\centering
\includegraphics[width=0.95\linewidth]{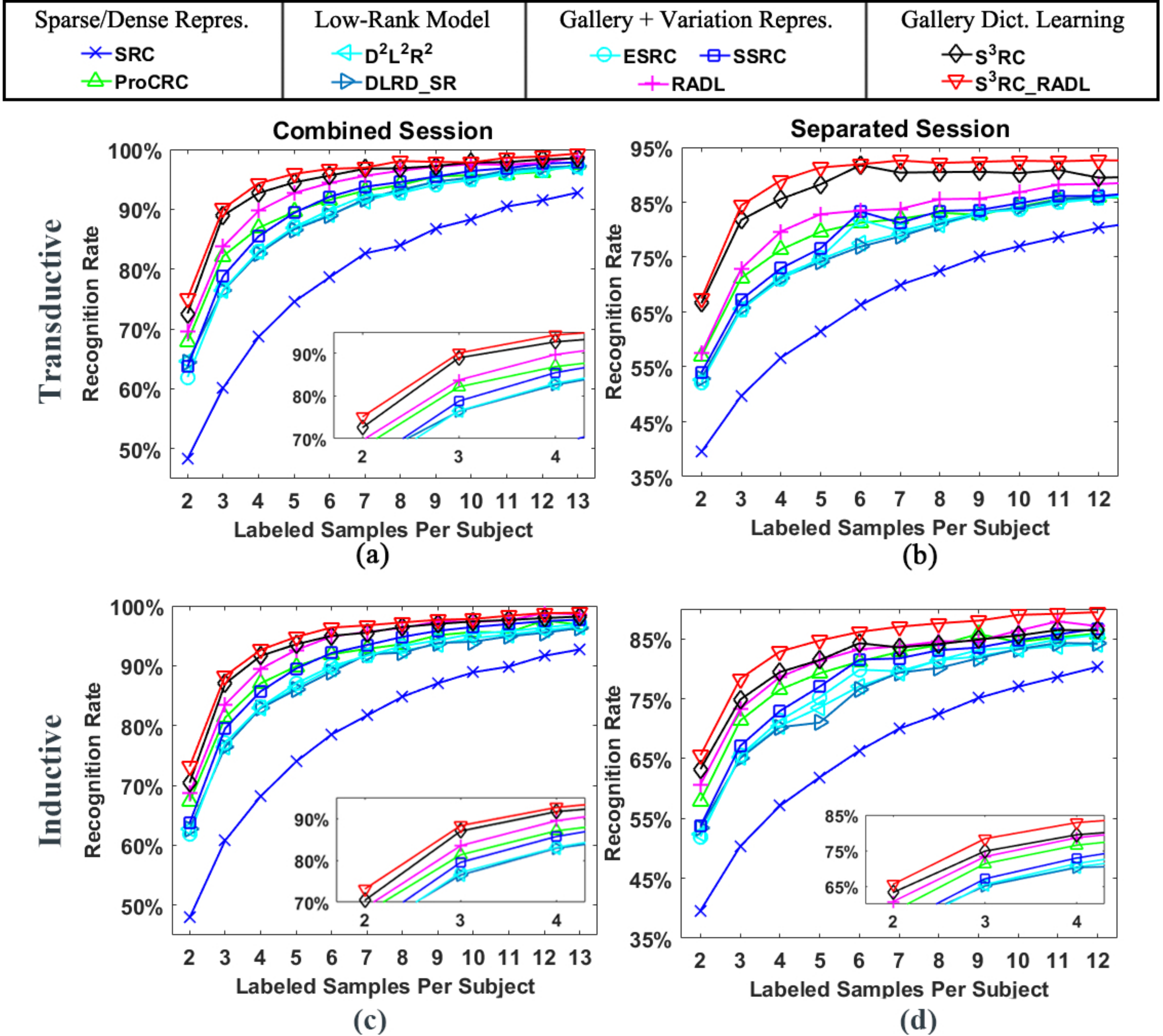}
\caption{The results (Recognition Rate) from the AR database. We used the first 300 PCs (dimensional reduction by PCA) and $\lambda$ is set to 0.005 (as identical of \cite{Deng13}). Each value was obtained from 20 runs. The Left and Right Columns denote experiments with Combined and Separated Session, the Top and Bottom Rows represent the \emph{transductive} and \emph{inductive} settings, respectively. Zoom-in figures are provided in subfigs (a), (c), and (d).}
\label{fig:AR_Results}
\end{figure}
Figure \ref{fig:AR_Results} shows that our results, \ie S$^3$RC and S$^3$RC-RADL, consistently outperform other counterparts under the same configurations. Moreover, significant improvements in the results are observed when there are few labeled samples per subject. Especially, it is observed that the results of SSRC are less satisfactory comparing with more recent state-of-the-art methods RADL and ProCRC, however, its performance is boosted to the second highest by utilizing the proposed gallery dictionary method S$^3$RC. For example, the accuracy of S$^3$RC is higher by around 10\% than SSRC when using 2-4 labeled samples per person. Furthermore, by combining with more recent RADL method, S$^3$RC-RADL achieves the best performance. It is also noted that the size of outperformance decreases when more labeled data is used. This is because the class centroids estimated by SSRC (averaging the labeled data according to same label) are less likely to be the true gallery when the number of labeled data are small. Thus, by improving the estimates of the true gallery from the initialization of the averaged labeled data, better results can be obtained by our method. Conversely, if the number of labeled data is sufficiently large, then the averaged labeled data becomes good estimates of the true gallery, which results in less improvement compared with our method.

The results of Figs. \ref{fig:AR_Results}a and \ref{fig:AR_Results}c (\ie the Left Column) are higher than those of Figs. \ref{fig:AR_Results}b and \ref{fig:AR_Results}d (\ie the Right Column) for all methods, because the labeled and unlabeled samples of the Right Column of Fig. \ref{fig:AR_Results} are obtained from different sessions. Interestingly, the higher outperformance of S$^3$RC can be observed in the more challenging experiment shown in Fig. \ref{fig:AR_Results}b. This observation further demonstrates the effectiveness of our proposed semi-supervised gallery dictionary learning method. Also, the results of the transductive experiments (\ie Figs. \ref{fig:AR_Results}a and \ref{fig:AR_Results}b) are better than those of the inductive experiments (\ie Figs. \ref{fig:AR_Results}c and \ref{fig:AR_Results}d), because the testing samples have been directly used to learn the model in the transductive settings.  


The above results have demonstrated the effectiveness of our method for the insufficient training samples problem.

\subsection{The performance on the SLSPP problem using facial image with neutral expression as gallery}
\subsubsection{The Multi-PIE Database \label{Sect:MultiPie}}

The large-scale Multi-PIE database \cite{Gross09} consists of images of four sessions (dates) with variations of pose, expression, and illumination. For each subject in each session, there are 20 illuminations, with indices from 0 to 19, per pose per expression. In our experiments, all the images are cropped to the size of 100 $\times$ 82. Since the data provider did not label the eye centers of each image in advance, we average the 4 labeled points of each eye ball (Points 38, 39, 41, 42 for the left eye and Points 44, 45, 47, 48 for the right eye) as the eye center, then crop them by locating the two eye centers at (19, 28) and (63, 28) of the 100 $\times$ 82 images. 

This experiment is a reproduction of the experiment on the SLSPP problem using the Multi-PIE database in \cite{Yang13}. Specifically, the images with illumination 7 from the first 100 subjects (among all 249 subjects) in Session 1 of the facial image with neutral expression (Session 1, Camera 051, Recording 1. $S1\_Ca051\_R1$ for short) are used as gallery. The remaining images under various illuminations of the other 149 subjects in $S1\_Ca051\_R1$ are used as the generic training data. For the testing data, we use the images of the first 100 subjects from other subsets of the Multi-PIE database, \ie the image subsets that are with different illuminations ($S2\_Ca051\_R1$, $S3\_Ca051\_R1$, $S4\_Ca051\_R1$), different illuminations and poses ($S1\_Ca050\_R1$, $S2\_Ca050\_R1$, $S3\_Ca050\_R1$, $S4\_Ca050\_R1$), different illuminations and expressions ($S1\_Ca051\_R2$, $S2\_Ca051\_R2$, $S2\_Ca051\_R3$, $S3\_Ca051\_R2$), different illuminations, expressions and poses ($S1\_Ca050\_R2$, $S1\_Ca140\_R2$).
The gallery image from a specific subject and its corresponding unlabeled/testing images with a randomly selected illumination are shown in Fig. \ref{fig:MPinstance}.
\begin{figure}[t]
\setcounter{figure}{5}
\centering
\includegraphics[width=0.9\linewidth]{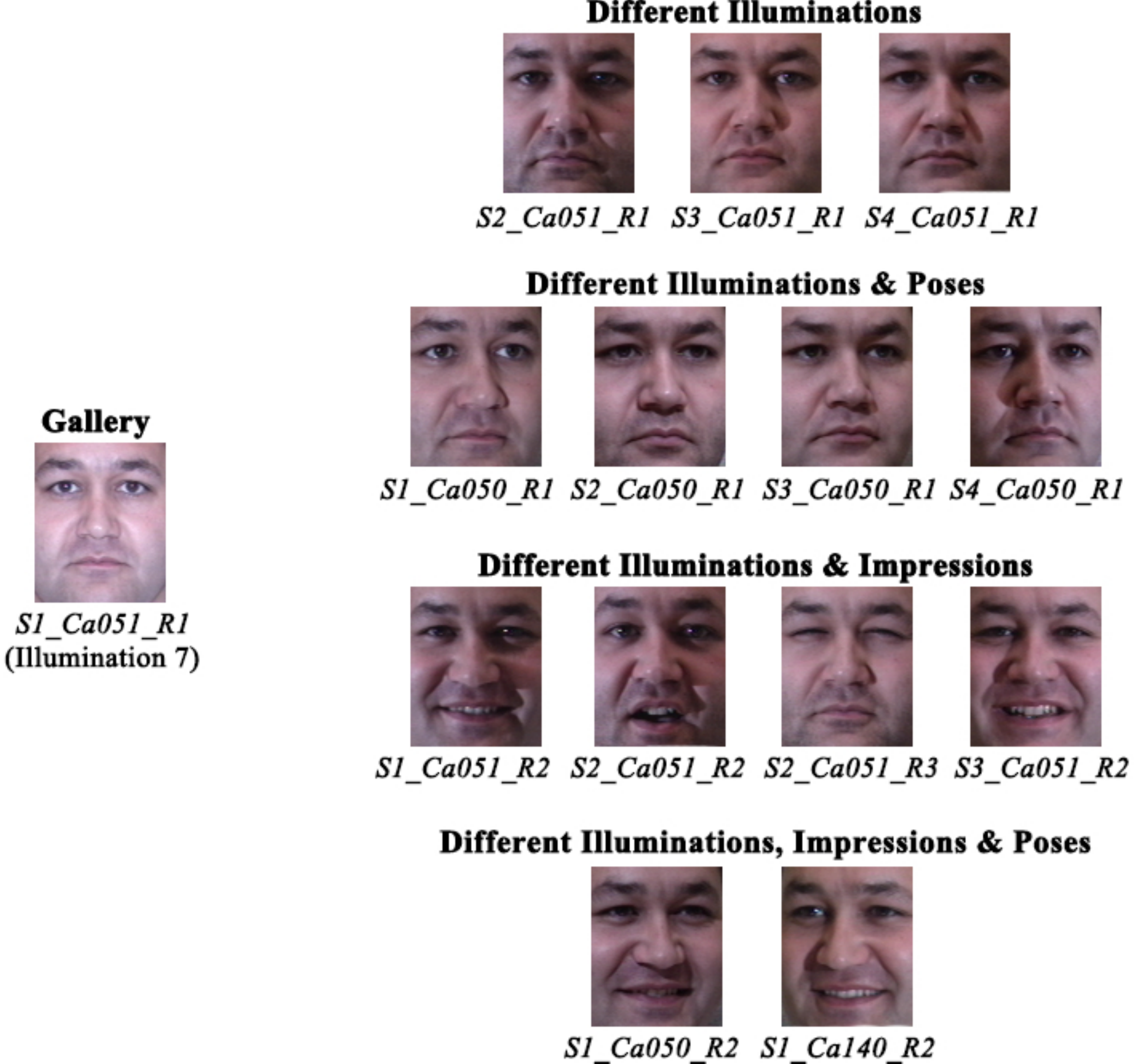}
\caption{The cropped image instances of one subject from various subsets of the Multi-PIE database.}
\label{fig:MPinstance}
\end{figure}

\begin{figure*}[t]
\setcounter{figure}{6}
\centering
\includegraphics[width=0.95\linewidth]{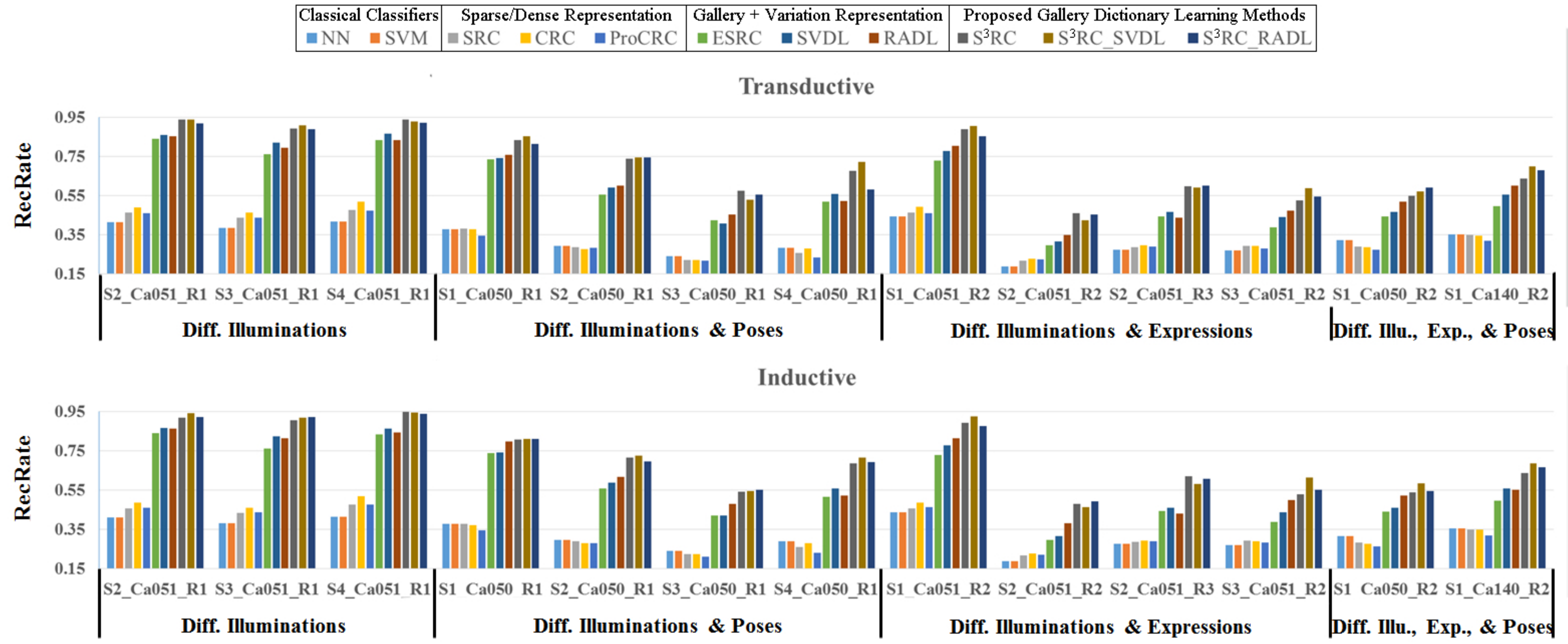}
\caption{The results (Recognition Rate, \%) from the Multi-PIE database with \textbf{controlled} single gallery sample per person, Top: RecRate for \emph{transductive} experiments, Bottom: RecRate for \emph{inductive} experiments. The bars from Left to Right are: NN, SVM, SRC, CRC, ProCRC, ESRC, SVDL, S$^3$RC, S$^3$RC-SVDL, and S$^3$RC-RADL.}
\label{fig:MultiPie}
\end{figure*}

\begin{figure*}[t]
\setcounter{figure}{7}
\centering
\includegraphics[width=0.85\linewidth]{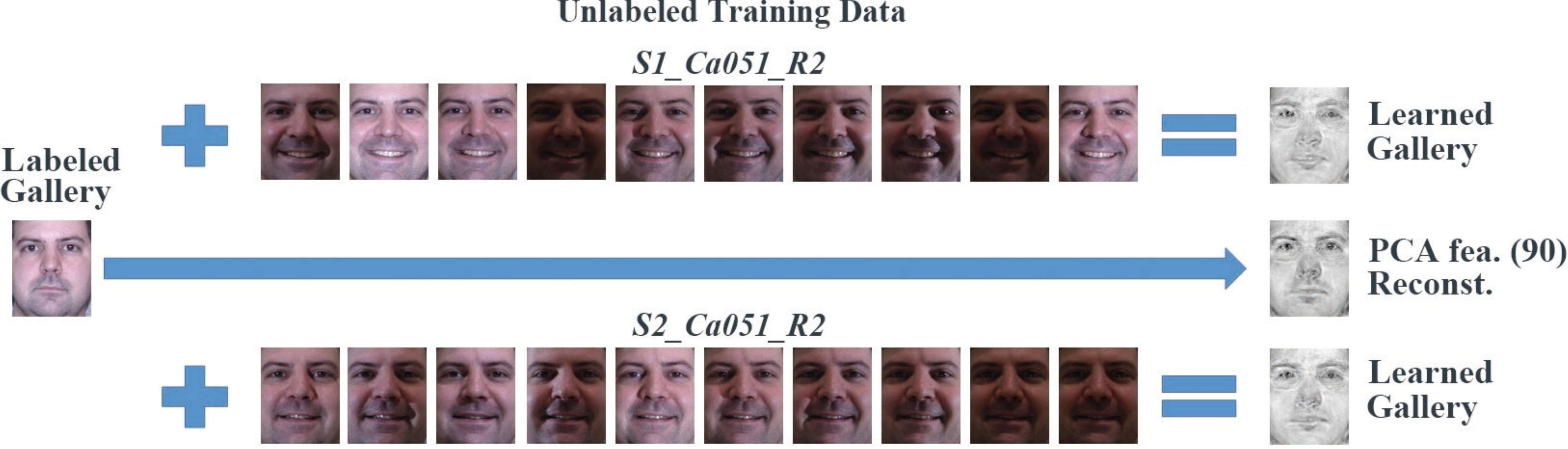}
\caption{Illustrations of the learned gallery samples when there is non-linear variation between the (input) labeled gallery and the unlabeled/testing samples. The labeled gallery is controlled facial image with neutral expression under standard illumination.}
\label{fig:Illu_PIE}
\end{figure*}

The results from classical classifiers NN \cite{Bishop06}, SVM \cite{Chang11}, sparse/dense representation SRC \cite{Wright09}, CRC \cite{Zhang11}, ProCRC \cite{Caiprobabilistic16}, gallery plus variation representation ESRC \cite{Deng12}, SVDL \cite{Yang13}, RADL \cite{Wei15}
are chosen for comparison\footnote{ Note that the DLRD\_SR \cite{Ma12} and $D^2 L^2 R^2$ \cite{Li13,Li14} methods, which we compared in the insufficient training samples problem, are less suitable for comparison here due to the SLSPP problem. It is because in order to learn a low-rank (sub-)dictionary for each subject, both of them assume low-rank property of the gallery dictionary, which requires multiple gallery samples per subject. SSRC also requires multiple gallery samples per subject to learn the gallery dictionary and thus is less suitable for comparison either.}
. In order to further investigate the generalizability of our method and to show the power of the gallery dictionary estimation, besides the evaluation on S$^3$RC-RADL, we also report the results of S$^3$RC using the variation dictionary learned by SVDL (S$^3$RC-SVDL), \ie initializing the first four steps of the Algorithm \ref{algorithm} by SVDL. The parameters were identical to those in \cite{Yang13}, \ie the 90 PCA dimension and $\lambda = 0.001$. The transductive and inductive experimental settings are:
\begin{itemize}
	\item \emph{Transductive}: Here we use all the images from the corresponding session as unlabeled testing data, the results are summarized in the top subfigure of Fig. \ref{fig:MultiPie}.
	\item \emph{Inductive}: The bottom subfigure of Fig. \ref{fig:MultiPie} summarizes the inductive experimental results, where the 20 images of the corresponding session were partitioned into two parts, \ie half for unlabeled training and the other half for testing. The inductive results are obtained by averaging 20 replicates.
\end{itemize}


Figure \ref{fig:MultiPie} shows that the proposed gallery dictionary learning methods, \ie S$^3$RC, S$^3$RC-SVDL, S$^3$RC-RADL achieve the top recognition rates compared with the other categories of methods in recognizing all the subsets. In particular, the highest enhancements can be observed for recognition with varying expressions. The reason might be that the samples with different expressions cannot be properly aligned by using only the eye centers, so the gallery dictionary learned by S$^3$RC can achieve better alignment with the testing images. This demonstrates that gallery dictionary learning plays an important role in SRC based face recognition. In addition, our inductive results are comparable to the transductive results, which implies we might do not need as many as 20 images as unlabeled training to learn the model on Multi-PIE database. It is also observed that in this challenging SLSPP problem, all the gallery plus variation representation methods (\ie ESRC, SVDL, RADL) outperform the best sparse/dense representation method (\ie ProCRC), suggesting that the gallery plus variation representation methods are more suitable for the SLSPP problem. As a generally applicable gallery dictionary learning method to the gallery plus variation framework, our method further improves the performance on this challenging tasks.

Furthermore, by integrating the variation dictionary learned by SVDL into S$^3$RC, S$^3$RC-SVDL, S$^3$RC-RADL also improves the performance of S$^3$RC in most cases. This also demonstrates the generalizablity of our method. These performance enhancements of S$^3$RC, S$^3$RC-SVDL and S$^3$RC-RADL (\emph{w.r.t} ESRC, SVDL, and RADL) are benefited from using the unlabeled samples for estimating the true gallery instead of relying on the labeled samples only. When given insufficient labeled samples, the other methods find it is hard to achieve satisfactory recognition rates in some cases.

The learned gallery is also investigated. We are especially interested in examining the learned gallery when there is a large difference between the input labeled samples and the unlabeled/testing images. Therefore, we use the neutral image as the labeled gallery and randomly choose 10 images with smile (\ie from $S1\_Ca051\_R2$ and $S2\_Ca051\_R2$), the learned gallery is shown in Fig. \ref{fig:Illu_PIE}. Figure \ref{fig:Illu_PIE} illustrates that by using the unlabeled training samples with smile, the learned gallery can also possess desirable (non-linear) smile attributes (see the mouth region), which better represents the prototype of the unlabeled/testing images. In fact, the proposed semi-supervised gallery dictionary learning method can be regarded as a pixel-wise alignment between the labeled gallery and the unlabeled/testing images.

The analysis in this section demonstrates the promising performance of our method for the SLSPP problem on Multi-PIE database.

\subsubsection{The CAS-PEAL Database}
The CAS-PEAL database \cite{Gao08} contains 99594 images with different illuminations, facing directions, expressions, accessories, etc. It is considered to be the largest database available that contains occluded images. Note that although the occlusions may commonly occurs on the objects of interests in practice \cite{Gao16,Gao17}, the Multi-PIE database does not contain images with occlusions. Thus, as a complementary experiment, we use all the 434 subjects from the \emph{Accessory} category for testing, and their corresponding images from the \emph{Normal} category as gallery. In this experimental setting, there are 1 neutral image, 3 images with hats, and 3 images with glasses/sunglasses for each subject. All the images are cropped to 100 $\times$ 82, with the centers of both eyes located at (19, 28) and (63, 28). Figure \ref{fig:CASinstance} illustrates the gallery and testing images for a specific subject.
\begin{figure}[t]
\centering
\includegraphics[width=0.9\linewidth]{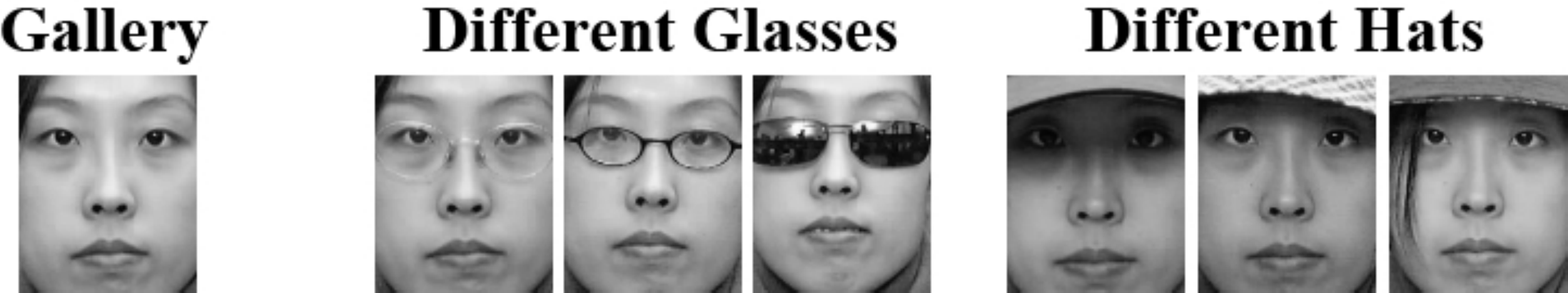}
\caption{The cropped image instances for one subject from the \emph{Normal} and the \emph{Accessory} categories of the CAS-PEAL database.}
\label{fig:CASinstance}
\end{figure}

Among the 434 subjects used, 300 subjects are selected for training and testing, and the remaining 134 subjects are used as generic training data. The first 100 dimension PCs (dimensional reduction by PCA) are used and $\lambda$ is set to 0.001. We also compare our results with NN \cite{Bishop06}, SVM \cite{Chang11}, SRC \cite{Wright09}, CRC \cite{Zhang11}, ProCRC \cite{Caiprobabilistic16}, ESRC \cite{Deng12}, SVDL \cite{Yang13} and RADL \cite{Wei15}. The results of the transductive and inductive experiments are reported in Table \ref{table:CAS}. The experimental settings are:
\begin{itemize}
	\item \emph{Transductive}: All the 6 images of the \emph{Accessory} category shown in Fig. \ref{fig:CASinstance} are used for unlabeled/testing.
	\item \emph{Inductive}: Of all the 6 images, we randomly select 3 images as unlabeled training and the remaining 3 images are used as testing. The inductive results are obtained by averaging 20 replicates.
\end{itemize}

Table \ref{table:CAS} shows that S$^3$RC, S$^3$RC-SVDL and S$^3$RC-RADL achieve top three recognition rates in the CAS-PEAL database, only except S$^3$RC-RADL vs. RADL in the \emph{Inductive} case. This is the only case that our method slightly inferior than our counterpart among all of our experiments (including the experiments in the previous and following sections). The reason is that in the inductive experiments, there is too few (i.e. 3 samples per subject) unlabeled data to guarantee the generalizibility of them. Also RADL used weighted $\ell_2$ norm (\ie $\ell_2$ norm with projection) to calculate the data error term, which already gains some robustness to the less perfect gallery dictionary. The results in Table \ref{table:CAS} verify the promising performance of our method for SLSPP problem on CAS-PEAL database.
\begin{table}[!htb]
\centering
\fontsize{8pt}{0.8\baselineskip}\selectfont
\caption{ The results (Recognition Rate, \%) from the CAS-PEAL database on the \emph{Normal} and \emph{Accessory} subsets, for both \emph{transductive} and \emph{inductive} experiments. In the brackets, we show the improvement of S$^3$RC \emph{w.r.t} ESRC, S$^3$RC-SVDL \emph{w.r.t} SVDL and S$^3$RC-RADL \emph{w.r.t} RADL, since the same variation dictionary has been used in these three pairs. We used the first 100 PCs (dimensional reduction by PCA) and $\lambda$ is set to 0.001.}
\begin{tabular}{@{\extracolsep{\fill}} c c c}\\
\toprule
Method  & Transductive & Inductive   \\
\midrule
NN 		& 41.00  & 41.08 \\
\cmidrule{1-3}
SVM 	& 41.00  & 41.08 \\
\cmidrule{1-3}
SRC     & 57.22       & 56.70             \\
\cmidrule{1-3}
CRC 	& 54.56  & 53.84 \\
\cmidrule{1-3}
ProCRC	& 54.89 	& 54.67 		\\
\cmidrule{1-3}
ESRC    & 71.78       & 71.76            \\
\cmidrule{1-3}
SVDL    & 69.67       & 69.74             \\
\cmidrule{1-3}
RADL  	& 75.27 	  & 73.78 					\\
\cmidrule{1-3}
S$^3$RC    & 75.39 ($\uparrow$3.61)       & {\bf 74.66} ($\uparrow$2.90)       \\
\cmidrule{1-3}
S$^3$RC-SVDL  & 72.06 ($\uparrow$2.39)     & 71.79 ($\uparrow$2.05)             \\
\cmidrule{1-3}
S$^3$RC-RADL  & {\bf 75.89} ($\uparrow$0.62)     & 72.56 ($\downarrow$1.23)  \\
\bottomrule
\end{tabular}
\label{table:CAS}
\end{table}

\subsection{The performance on the SLSPP problem using uncontrolled image as gallery}
\subsubsection{The Multi-PIE Database \label{sect:multi-PIE-unconstrained}}
In order to validate the proposed S$^3$RC methods, additional more challenging experiments are performed on the Multi-PIE Database, where an uncontrolled image is used as the labeled gallery. Specifically, for each unlabeled/testing subset illustrated in Fig. \ref{fig:Uncontrol_MultiPIE}, we randomly choose one image per subject from the other subsets (excluding the unlabeled/testing subset) as the labeled gallery. It should be noted that the well controlled gallery, \ie the neutral images from $S1\_Ca050\_R1$, is not used in this section. Both the transductive and the inductive experiments are also reported as the same protocol used in Sect. \ref{Sect:MultiPie}.

\begin{figure}[!htp]
\setcounter{figure}{9}
\centering
\includegraphics[width=0.9\linewidth]{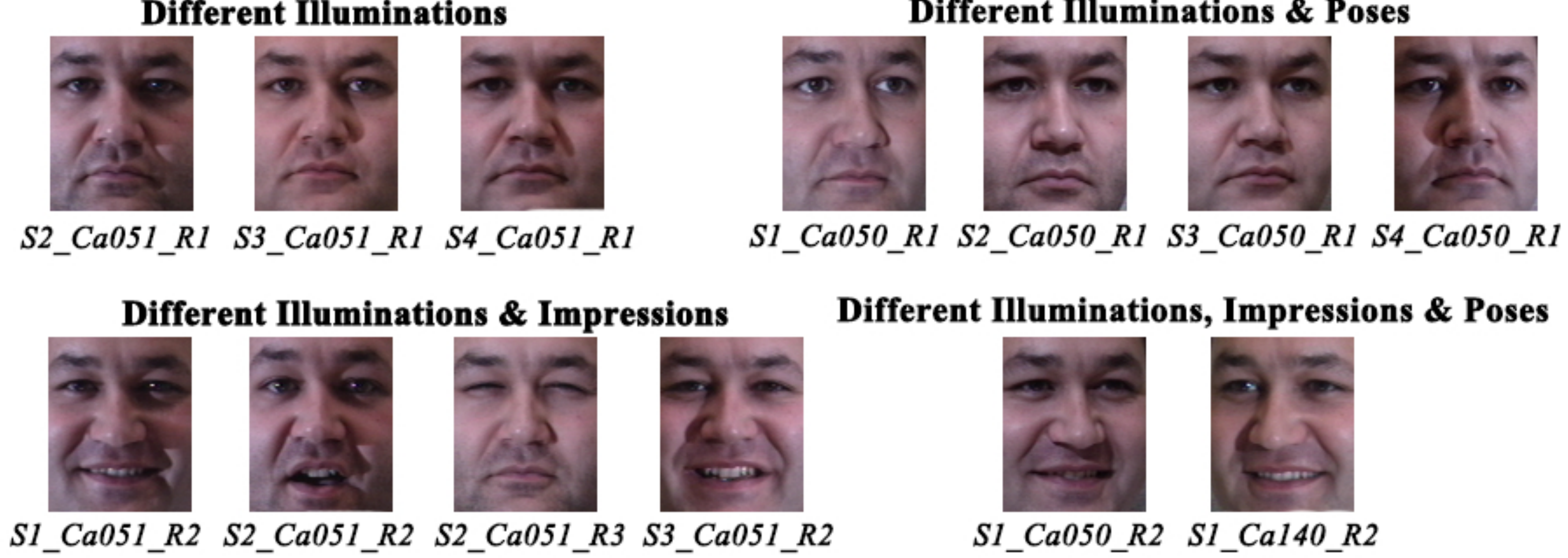}
\caption{The illustrations of the data used in Sect. \ref{sect:multi-PIE-unconstrained}. We first randomly select a subset as unlabeled/testing data, then a gallery sample is randomly chosen from other subsets (excluding the unlabeled/testing select).}
\label{fig:Uncontrol_MultiPIE}
\end{figure}

\begin{figure*}[t]
\setcounter{figure}{10}
\centering
\includegraphics[width=0.95\linewidth]{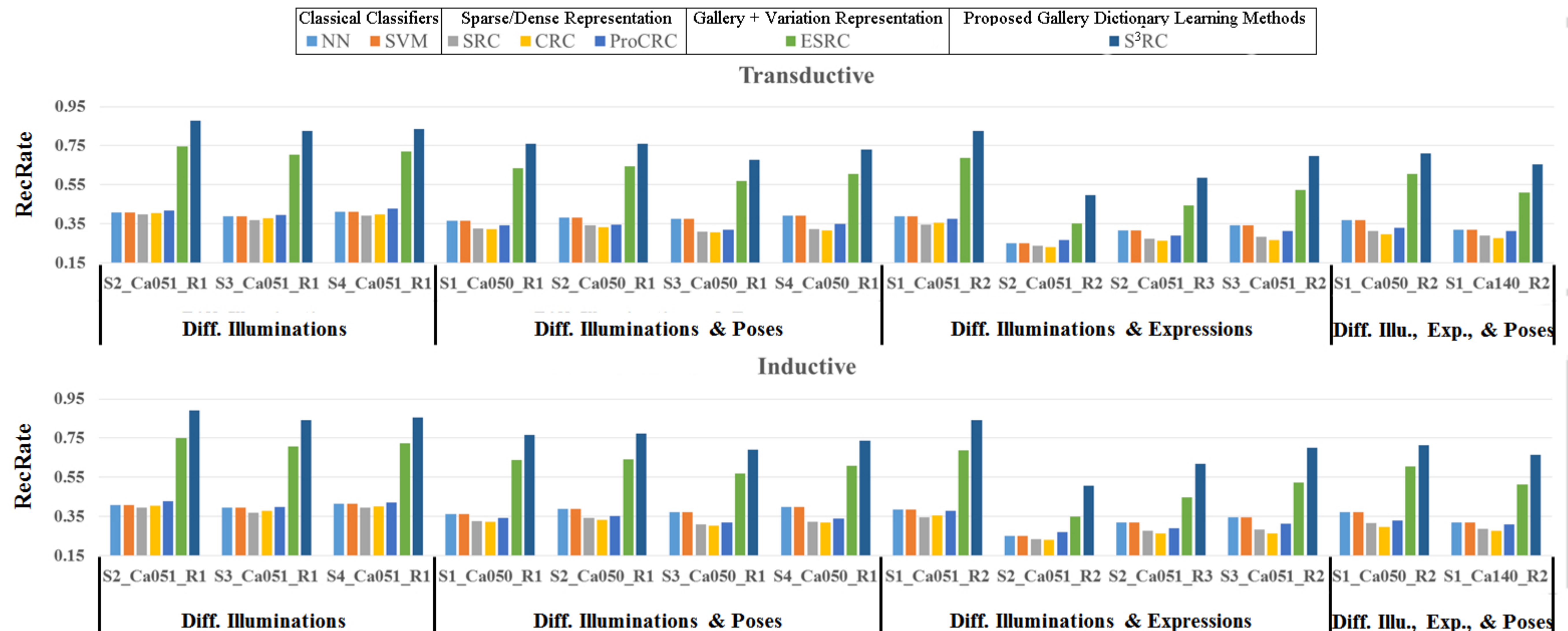}
\caption{The results (Recognition Rate, \%) from the Multi-PIE database with \textbf{uncontrolled} single gallery sample per person, Top: RecRate for \emph{transductive} experiments, Bottom: RecRate for \emph{inductive} experiments. The bars from Left to Right are: NN, SVM, SRC, CRC, ProCRC, ESRC, S$^3$RC.}
\label{fig:MultiPie_rand}
\end{figure*}

\begin{itemize}
	\item \emph{Transductive}: Here we use all the images from the corresponding session as unlabeled testing data, the results are summarized in top subfigure of Fig. \ref{fig:MultiPie_rand}.
	\item \emph{Inductive}: Bottom subfigure of Fig. \ref{fig:MultiPie_rand} summarizes the inductive experimental results, where the 20 images of the corresponding session are partitioned into two parts, \ie half for unlabeled training and the other half for testing. The inductive results are obtained by averaging 20 replicates.
\end{itemize}

\begin{figure*}[t]
\setcounter{figure}{11}
\centering
\includegraphics[width=0.85\linewidth]{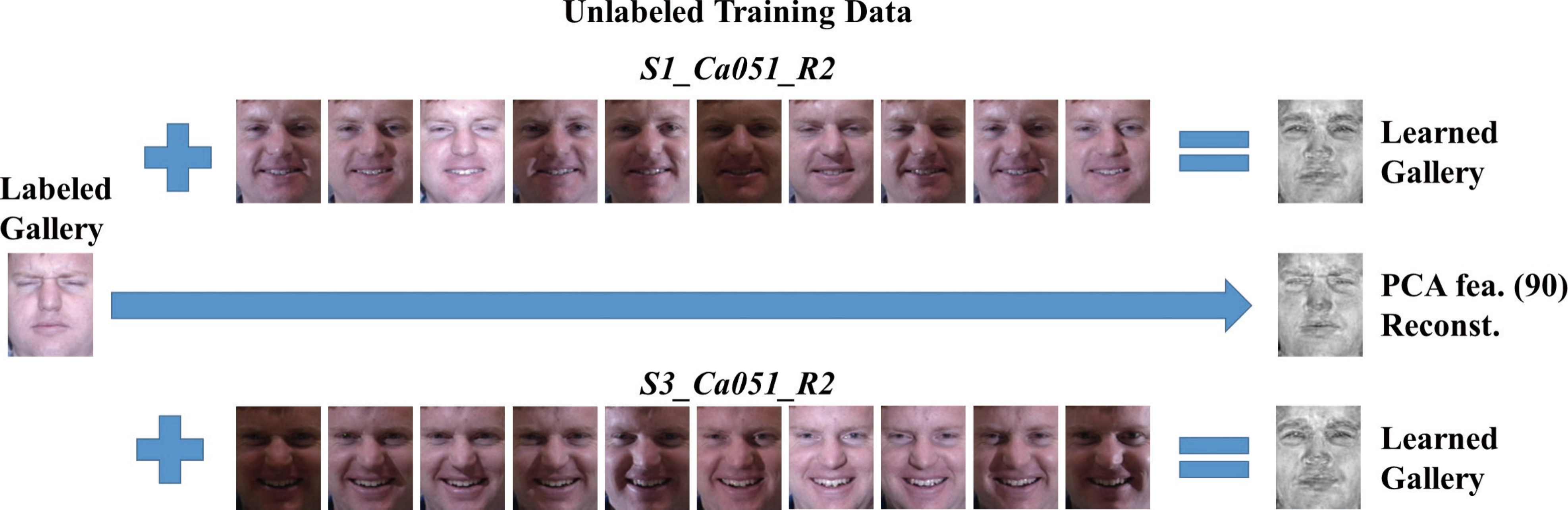}
\caption{Illustrations of the learned gallery samples when there is non-linear variation between the (input) labeled gallery and the unlabeled/testing samples. The labeled gallery is uncontrolled (\ie the squint image).}
\label{fig:Illu_PIE2}
\end{figure*}

The results are shown in Fig. \ref{fig:MultiPie_rand}, in which the proposed S$^3$RC is compared with NN, SVM, SRC \cite{Wright09}, CRC, and ProCRC methods\footnote{Note that the SVDL and RADL method is less suitable to the SLSPP problem with uncontrolled image as gallery. It is because the variation dictionary learning of SVDL or RADL requires reference images for all the subjects in the generic training data, where the reference images should have the same type of variation as the gallery. However, such information cannot be inferred due to the uncontrolled gallery images used.}. Figure \ref{fig:MultiPie_rand} shows that our method consistently outperforms the other outline methods. In fact, although the overall accuracy decreases due to the uncontrolled labeled gallery, all the conclusions made in Sect. \ref{Sect:MultiPie} are supported and verified by Fig. \ref{fig:MultiPie_rand} here.

We also investigated the learned gallery when a uncontrolled image is used as input labeled gallery. Figure \ref{fig:Illu_PIE2} shows the results when using the squint image (\ie a image from $S2\_Ca051\_R3$) as the single labeled input gallery for each subject. It is observed (see eye and mouth regions) that the learned gallery can better represent the testing images (\ie smile) by the non-linear semi-supervised gallery dictionary learning. The reason is the same as the previous experiments in Fig. \ref{fig:Illu_PIE}, \ie the proposed semi-supervised method conducts a pixel-wise alignment between the labeled gallery and the unlabeled/testing images so that the non-linear variations between them are well addressed.

\subsubsection{The LFW Database}
The Labeled Face in the Wild (LFW) \cite{LFWTech} is the latest benchmark database for face recognition, which has been used to test several advanced methods with dense or deep features for face verification, such as \cite{Parkhi15,sun2014deep,sun2015deeply,chen2013blessing,taigman2014deepface}. In this section, our method has been tested on the LFW database to create a more challenging face identification problem.

Specifically, the face images of the LFW database are collected from the Internet, as long as they can be detected by the Viola-Jones face detector \cite{LFWTech}. As a result, there are more than 13,000 images in the LFW database containing enormous intra-class variations, where controlled (\eg neutral and facial) faces may not be available. Considering the previous experiments on the Multi-PIE and CAS-PEAL databases dealt with specific kinds of variations separately, as an important extended experiment, the effectiveness of our S$^3$RC method can be further validated by its performance on the LFW dataset.

A pre-aligned database by deep funneling \cite{Huang2012a} was used in our experiment. We select a subset of the LFW database containing more than 10 images for each person, with 4324 images from 158 subjects in total. In the experiments, we randomly select 100 subjects for training and testing the model, the remaining 58 subjects are used to construct the variation dictionary. The experimental results for both transductive and inductive learning are reported. The only gallery image is randomly chosen from each subject, then the transductive and the inductive experimental settings are:
\begin{itemize}
	\item \emph{Transductive}: All the remaining images from a specific subject are used for unlabeled/testing. The results are obtained by averaging 20 replicates due to the randomly selected subjects.
	\item \emph{Inductive}: For each subject, we randomly select half of the remaining images as unlabeled training and the other half are used for testing. The results are obtained by averaging 50 replicates due to the randomly selected subjects and random unlabeled-testing split.
\end{itemize}

The results are shown in Table \ref{table:LFW}, where we use NN, SVM, SRC \cite{Wright09}, CRC, and ProCRC for comparison. First, we have tested our method using simple features obtained from an unsupervised dimensional reduction by PCA. The results in the left two columns of Table \ref{table:LFW} show that, although none of the methods achieve a satisfactory performance, our method, based on the semi-supervised gallery dictionary, still significantly outperforms the baseline methods.

Nowadays, Deep Convolution Neural Network (DCNN) based methods have achieved state-of-the-art performance on the LFW database \cite{Parkhi15,sun2014deep,sun2015deeply,chen2013blessing,taigman2014deepface}. It is noticed that the DCNN methods often use basic classifiers to do the classification, such as softmax, linear SVM or $\ell_2$ distance. Recently, it is showed that by coupling with the deep-learned CNN features, the SRC methods can achieve significantly improved results \cite{Caiprobabilistic16}. Motivated by this, we also aim to verify that by utilizing the same deep-learned features, our method (\ie our classifier) is able to further improve the results obtained by the basic classifiers.

Specifically, we utilize a recent and public DCNN model named VGG-face \cite{Parkhi15} to extract the 4096-dimensional features, then our method, as well as the baseline methods, are implied to perform the classification. The results, shown in the left two-column of Table \ref{table:LFW}, demonstrate the significantly improved results from the proposed methods using the DCNN features, whereas such an investigation cannot be observed by comparing other SRC methods, \eg SRC, CRC, ESRC, with the basic NN classifier. It is noted that the original classifier used in \cite{Parkhi15} is the $\ell_2$ distance in face verification, which is equivalent to KNN ($K = 1$) in face identification with SLSPP. Therefore, the results in Table \ref{table:LFW} demonstrate that with the state-of-the-art DCNN features, the performance on the LFW database can be further boosted by using the proposed semi-supervised gallery dictionary learning method.

\begin{table}[!htb]
\centering
\fontsize{8pt}{0.8\baselineskip}\selectfont
\caption{ The results (Recognition Rate, \%) from the LFW database, for both \emph{transductive} and \emph{inductive} experiments with simple PCA features and deep learned feature by \cite{Parkhi15}. In the brackets, we show the improvement of S$^3$RC \emph{w.r.t} ESRC. We used the first 100 PCs (dimensional reduction by PCA) and the deep learned features by \cite{Parkhi15} is of 4096 dimensions. $\lambda$ is set to 0.001.}
\begin{tabular}{@{\extracolsep{\fill}} c c c c c}\\
\toprule
\multirow{2}{*}{Method}  & \multicolumn{2}{c}{PCA fea. (100)} & \multicolumn{2}{c}{DCNN fea. by \cite{Parkhi15} (4096)} \\
\cmidrule(r{0.3em}){2-3} \cmidrule(l{0.3em}){4-5}
  & Transductive & Inductive & Transductive & Inductive   \\
\midrule
NN 		& 5.57			& 5.82       	&  89.28		& 90.19\\
\cmidrule{1-5}
SVM 	& 5.37			& 5.82 			&  89.28	    & 90.19\\
\cmidrule{1-5}
SRC     & 10.92       	& 11.13 		&  89.50		& 90.23       \\
\cmidrule{1-5}
CRC 	& 10.47			& 10.69 		&  89.18	    & 89.86 \\
\cmidrule{1-5}
ProCRC  & 10.77 		& 10.99 		&  90.85 		& 90.13 \\
\cmidrule{1-5}
ESRC    & 15.51       	& 16.23    		&  90.58		& 90.73        \\
\cmidrule{1-5}
S$^3$RC    & {\bf 17.99} ($\uparrow$2.48)   & {\bf 17.90} ($\uparrow$1.67)	& 	{\bf 92.55} ($\uparrow$1.98)	& {\bf 92.57} ($\uparrow$1.84)    \\
\bottomrule
\end{tabular}
\label{table:LFW}
\end{table}

\begin{figure}[!htp]
\centering
\includegraphics[width=\linewidth]{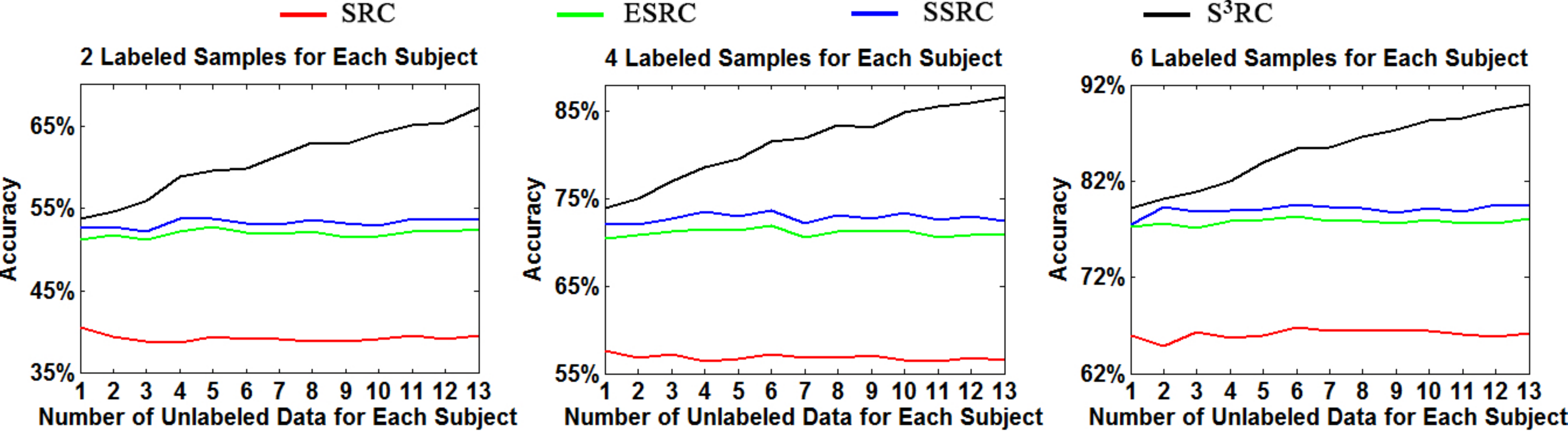}
\caption{The analysis of the impact of different amounts of unlabeled data from 1-13 on AR database. Three different amounts of labeled data are chosen for analysis, \emph{i.e.} 2, 4 and 6 labeled samples per subject, as shown in Left, Middle, Right subfigures, respectively. The results are obtained by averaging 20 runs, number of PCs is 300 (dimensional reduction by PCA) and $\lambda$ is 0.005.}
\label{fig:amount}
\end{figure}

\begin{figure*}[!htp]
\centering
\subfigure[]{
\label{fig:Alignment_a}
\includegraphics[width=0.265\linewidth]{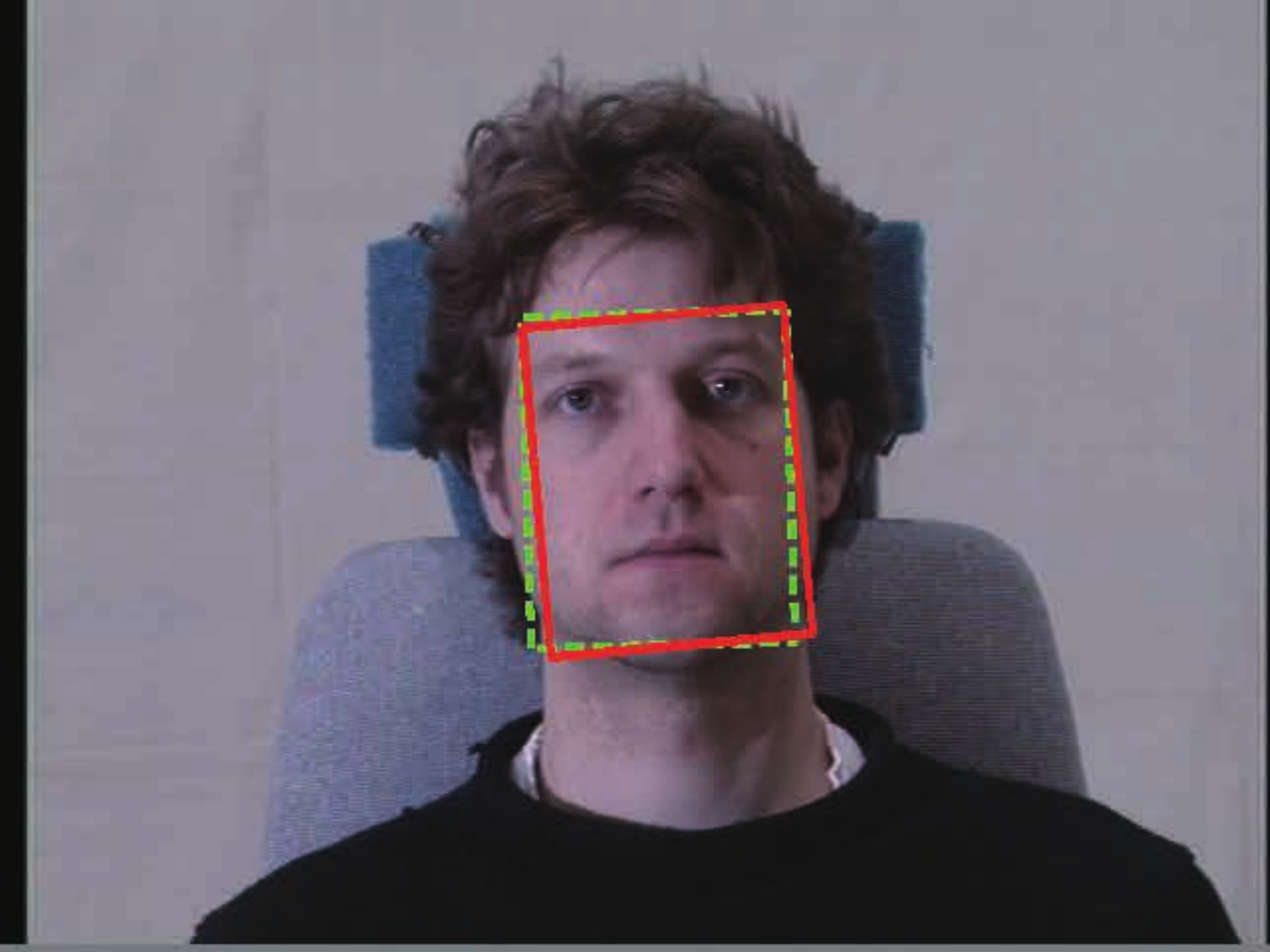}}
\hspace{0.4in}
\subfigure[]{
\label{fig:Alignment_b}
\includegraphics[width=0.2\linewidth]{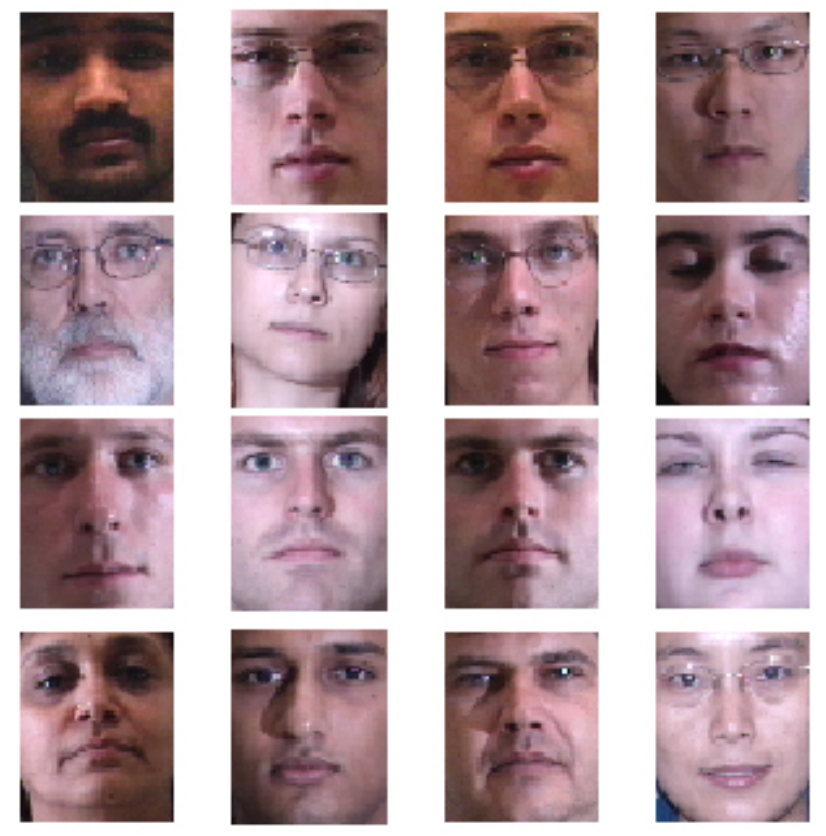}}
\hspace{0.4in}
\subfigure[]{
\label{fig:Alignment_c}
\includegraphics[width=0.2\linewidth]{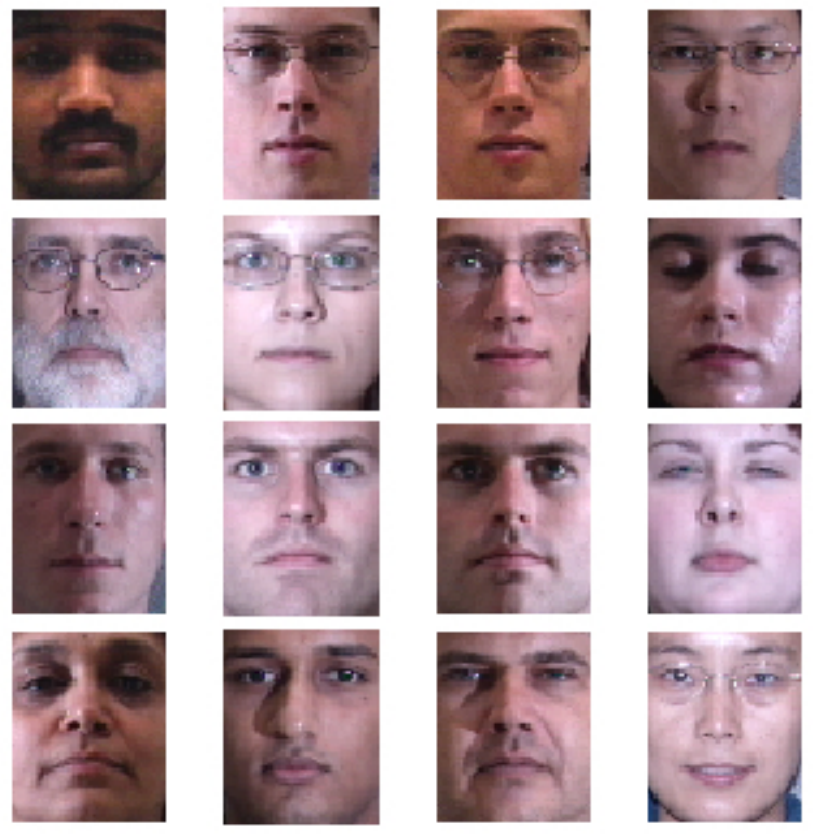}}

\caption{The examples of detected and aligned faces. (a) An example of detected faces (by Viola-Jones detector, green dash box) and aligned faces (by MRR, red solid box) on the original image. (b) The cropped detected faces. (c) The cropped aligned faces.}
\label{fig:Alignment}
\end{figure*}

\subsection{Analysis of the influence of different amounts of labeled and/or unlabeled data}
The impact of different amounts of unlabeled data in S$^3$RC is analyzed on different amounts of labeled data using AR database. In this experiment, we first randomly choose a session, and then select 1-13 unlabeled data for each subject from that session to investigate the influence of different amounts of unlabeled data. 2, 4 and 6 labeled samples of each subject are randomly chosen from the other session. For comparison, we also illustrate the results of SRC, ESRC, SSRC with the same configurations. The results are shown in Fig. \ref{fig:amount}, which is obtained by averaging 20 runs.

It can be observed from Fig. \ref{fig:amount} that: i) our method is effective (\emph{i.e.} can outperform the state-of-the-art) even when there is only 1 unlabeled sample per subject; ii) when more unlabeled samples are used, we observe significant increased accuracy from our method, while the accuracies of the state-of-the-art methods do not change much, because unlabeled information is not considered in these methods; iii) the better performance of our method compared with the alternatives is not affected by different amounts of labeled data used.

Furthermore, we are also interested in illustrating the learned galleries. Compared with the experiments on the AR database stated above, the experiments similar to those on the Multi-PIE database in Sect.\ref{Sect:MultiPie} are more suitable to our purpose. It is because in the above AR experiments, the influence of randomly selected gallery and unlabeled/testing samples can be eliminated by averaging the RecRate of multiple replicates. However, the learned galleries from multiple replicates cannot be averaged for illustration. Therefore, the only (fixed) labeled gallery sample per subject and the similarity between the unlabeled/testing samples in the Multi-PIE database enable to alleviate such influence. In addition, the large difference between the labeled gallery and the unlabeled/testing samples is more suitable to illustrate the effectiveness of the proposed semi-supervised gallery learning method.

\begin{figure}[!pt]
\centering
\includegraphics[width=0.9\linewidth]{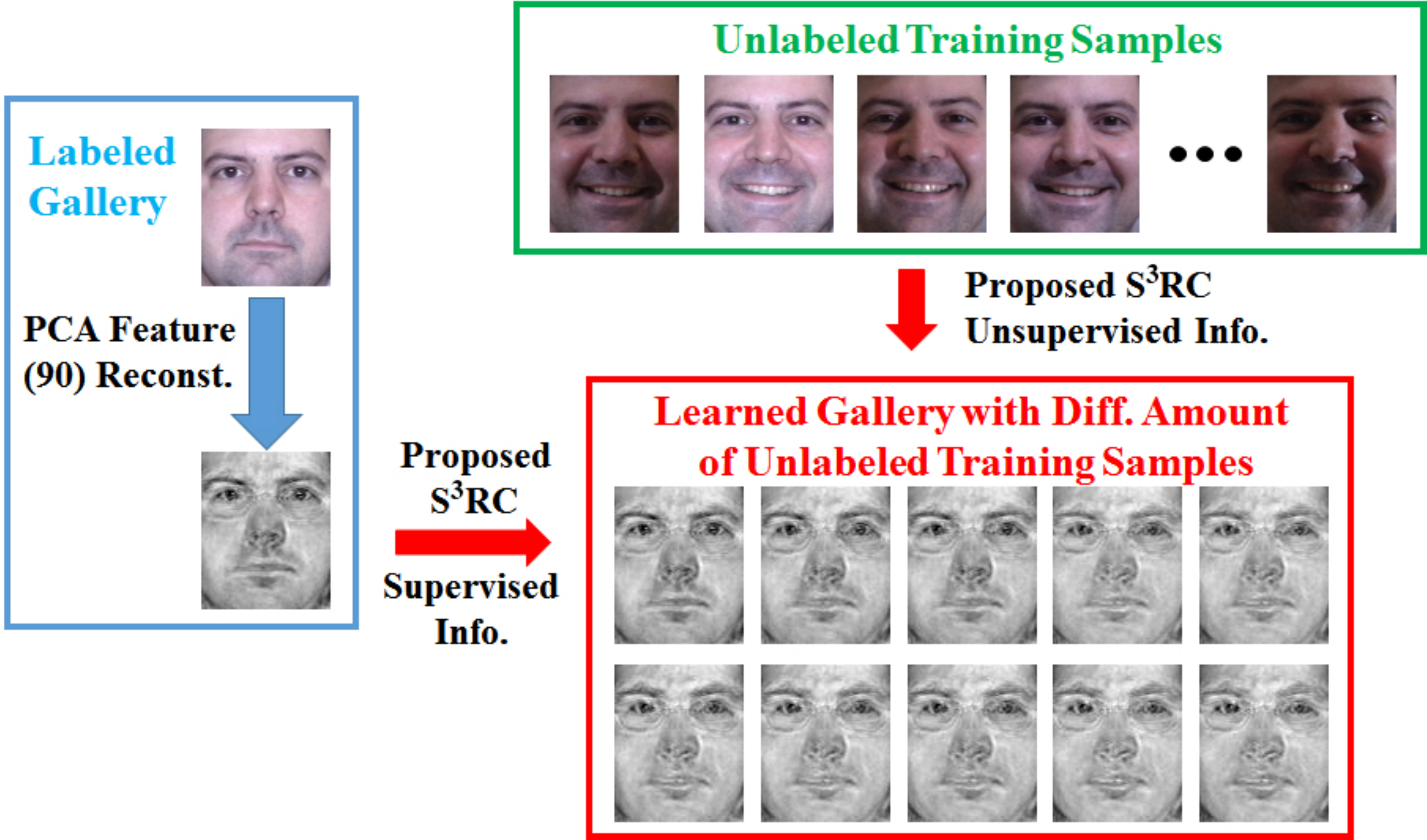}
\caption{The illustrations of the learned gallery with different amount (1-10 per subject) of unlabeled training samples. The reconstructed learned galleries are illustrated in the Red Box, with 1-10 unlabeled samples per subject from Left to Right, then Top to Bottom. Better representation can be observed with more unlabeled training samples (\ie smile, see the mouth region).}
\label{fig:AmountIllu}
\end{figure}

Specifically, the same neutral image from Sect.\ref{Sect:MultiPie} is used as the only labeled gallery sample per subject. Images from $S1\_Ca051\_R2$ are used as the unlabeled/testing images. We randomly choose 1-10 testing images to do the experiments, each trail is used to draw the learned galleries as each subfigure of Fig. \ref{fig:AmountIllu}. 

Figure \ref{fig:AmountIllu} shows that with more unlabeled training samples, the gallery samples learned by our proposed S$^3$RC method can better represent the unlabeled data (\ie \emph{smile}, see the mouth region). In fact, we note that the proposed semi-supervised learning method can be regarded as \emph{nonlinear pixel-wise/local alignments}, \eg to align the neutral gallery and the smiling unlabeled/testing faces in Fig. \ref{fig:AmountIllu}, therefore enabling a better representation of the unlabeled/testing data to achieve improved performance over the existing linear and global SRC methods (\eg SRC, ESRC, SSRC, \etc).

\subsection{The performance of our method with different alignments}
In the previous experiments, the performance of our method is investigated using face images that are aligned manually by eye centers. Here, we show the results using different alignments in a fully automatic pipeline. That is, for a given image, we use the Viola-Jones detector \cite{Viola04} for face detection, Misalignment-Robust Representation (MRR) \cite{Yang12} for face alignment, and the proposed S$^3$RC for recognition.

Note that the aim of this section is to prove that i) the better performance of S$^3$RC is not affected by different alignments, and ii) S$^3$RC can be integrated into a fully automatic pipeline for practical use. MRR is chosen for alignment, because the code is available online\footnote{MRR codes can be downloaded at \url{http://www4.comp.polyu.edu.hk/~cslzhang/code/MRR_eccv12.zip}.}. Researchers can also use other alignment techniques (\emph{e.g.} SILT \cite{Zhuang14, Zhuang13}, DSRC \cite{Wagner12} by aligning the gallery first, then align the query data to the well aligned gallery; or TIPCA \cite{Deng14}), but the comparison of different alignment methods is beyond the scope of this paper.

For MRR alignment, we use the default settings of the demo codes, and only change the input data. That is, the well aligned neutral images of the first 50 subjects of Multi-PIE database (as provided in the MRR demo codes) are used as the gallery to align the whole Multi-PIE database of 337 subjects, $\lambda$ of MRR is set to 0.05, the number of selected candidates is set to 8 and the output aligned data are cropped into 60 $\times$ 48 images. An example of detected and aligned result for a specific subject on the original image is shown in Fig. \ref{fig:Alignment_a}. Other detected and aligned face examples are shown in Figs. \ref{fig:Alignment_b} and \ref{fig:Alignment_c}, respectively.

The classification results using Viola-Jones for detection and MRR for alignment are shown in Table \ref{table:MultiPie3} for some subsets of Multi-PIE database with transductive experimental settings. The classification parameters are identical with them in Sections 3.2. Table \ref{table:MultiPie3} clearly shows that S$^3$RC and S$^3$RC-SVDL (\emph{i.e.} S$^3$RC plus the variation dictionary learned by SVDL) achieve the two highest accuracies. By using the same aligned data as input for all the methods, we show that the strong performance of S$^3$RC is due to the utilization of the unlabeled information, no matter whether the alignment is obtained manually or by automatic MRR.
\begin{table}[!htb]
\centering
\fontsize{8pt}{0.8\baselineskip}\selectfont
\caption{The results (Recognition Rate, \%) for the Multi-PIE database using Viola-Jones for detection and MRR for alignment. In the brackets, we show the improvement of S$^3$RC \emph{w.r.t} ESRC and S$^3$RC-SVDL \emph{w.r.t} SVDL, since the same variation dictionary has been used in both pairs. We used the first 90 PCs (dimensional reduction by PCA) and $\lambda$ is set to 0.001.}
\begin{tabular}{@{\extracolsep{\fill}} c c c c c c c c c}\\
\toprule
Method  & $S2\_Ca051\_R1$ & $S3\_Ca051\_R1$ & $S4\_Ca051\_R1$   \\
\midrule
SRC     & 55.75       & 51.47       & 53.64             \\
\cmidrule(r{-0.1em}){1-4}
ESRC    & 86.78       & 85.07       & 86.17             \\
\cmidrule(r{-0.1em}){1-4}
SVDL    & 91.78       & 89.30       & 89.68             \\
\cmidrule(r{-0.1em}){1-4}
S$^3$RC    & \underline{95.75} ($\uparrow$8.97)      & \underline{94.58} ($\uparrow$9.51)      & \underline{93.96} ($\uparrow$7.79)      \\
\cmidrule(r{-0.1em}){1-4}
S$^3$RC-SVDL  & {\bf 97.74} ($\uparrow$5.96)    & {\bf 95.28} ($\uparrow$5.98)   & {\bf 95.32} ($\uparrow$5.64)            \\
\bottomrule
\end{tabular}
\label{table:MultiPie3}
\end{table}

\section{Discussion}
A semi-supervised sparse representation based classification method is proposed in this paper. By exploiting the unlabeled data, it can well address both linear and non-linear variations between the labeled/training and unlabeled/testing samples. This is particularly useful when the amount of labeled data is limited. Specifically, we first use the gallery plus variation model to estimate the rectified unlabeled samples excluding linear variations. After that, the rectified labeled and unlabeled samples are used to learn a GMM using EM clustering to address the non-linear variations between them. These rectified labeled data is also used as the initial mean of the Gaussians in the EM optimization. Finally, the query samples are classified by SRC with the precise gallery dictionary estimated by EM.

The proposed method is flexible, in which the gallery dictionary learning method is complementary to existing methods which focus on learning the (linear) variation dictionary, such as ESRC \cite{Deng12}, SVDL \cite{Yang13}, SILT \cite{Zhuang13,Zhuang14}, RADL \cite{Wei15} \etc For the ESRC, SVDL and RADL methods, we have shown by experiments that the combination of the proposed gallery dictionary learning and ESRC (namely S$^3$RC), SVDL (namely S$^3$RC-SVDL) and RADL (namely S$^3$RC-RADL) achieve significantly improved performance on all the tasks.

It is also noted by coupling with state-of-the-art DCNN features, our method, as a better classifier, can further improve the recognition performance. This has been verified by using VGG-face \cite{Parkhi15} features on LFW database, where our method outperforms the other baselines by 1.98\%. While less improvement is observed by feeding DCNN feature to other classifiers, including SRC, CRC, ProCRC and ESRC, when comparing with the basic nearest neighbor classifier.


Moreover, our method can be combined with SRC methods that incorporate auto-alignment, such as SILT \cite{Zhuang14,Zhuang13}, MRR \cite{Yang12}, and DSRC \cite{Wagner12}. Note that these methods will degrade to SRC after the alignment (except SILT, while by which the learned illumination dictionary can also be utilized in S$^3$RC by the same approach as S$^3$RC-SVDL). Thus, these methods can be used to align the images first, then S$^3$RC can be applied for classification utilizing the unlabeled information. 
In practical face recognition tasks, our method can be used following on automatic pipeline of face detection (\eg Viola-Jones detector \cite{Viola04}), followed by alignment (\eg one of \cite{Deng14,Wagner12,Yang12,Zhuang14,Zhuang13,ma2016infrared,ma2015robust2}), and then S$^3$RC classification
. 

\section{Conclusion}
In this paper, we propose a semi-supervised gallery dictionary learning method called S$^3$RC, which improves the SRC based face recognition by modeling both linear and non-linear variation between the labeled/training and unlabeled/testing samples, and leveraging the unlabeled data to learn a more precise gallery dictionary. These better characterize the discriminative features of each subject. Through extensive simulations, we can draw the following conclusions: i) S$^3$RC can deliver significantly improved results for both the insufficient training samples problem and the SLSPP problems. ii) Our method can be combined with state-of-the-art method that focus on learning the (linear) variation dictionary, so that we can obtain further improved the results (\eg see ESRC v.s. S$^3$RC, SVDL v.s. S$^3$RC-SVDL, and RADL v.s. S$^3$RC-RADL). iii) The promising performance of S$^3$RC is robust to different face alignment methods. A future direction is to use SSL methods other than GMM to better estimate the gallery dictionary.





\ifCLASSOPTIONcaptionsoff
  \newpage
\fi



%


\bibliographystyle{IEEEtran}
\bibliography{S3RC}

\begin{thebibliography}{10}
\providecommand{\url}[1]{#1}
\csname url@samestyle\endcsname
\providecommand{\newblock}{\relax}
\providecommand{\bibinfo}[2]{#2}
\providecommand{\BIBentrySTDinterwordspacing}{\spaceskip=0pt\relax}
\providecommand{\BIBentryALTinterwordstretchfactor}{4}
\providecommand{\BIBentryALTinterwordspacing}{\spaceskip=\fontdimen2\font plus
\BIBentryALTinterwordstretchfactor\fontdimen3\font minus
  \fontdimen4\font\relax}
\providecommand{\BIBforeignlanguage}[2]{{%
\expandafter\ifx\csname l@#1\endcsname\relax
\typeout{** WARNING: IEEEtran.bst: No hyphenation pattern has been}%
\typeout{** loaded for the language `#1'. Using the pattern for}%
\typeout{** the default language instead.}%
\else
\language=\csname l@#1\endcsname
\fi
#2}}
\providecommand{\BIBdecl}{\relax}
\BIBdecl

\bibitem{Wright09}
J.~Wright, A.~Y. Yang, A.~Ganesh, S.~S. Sastry, and Y.~Ma, ``Robust face
  recognition via sparse representation,'' \emph{IEEE Transactions on Pattern
  Analysis and Machine Intelligence}, vol.~31, no.~2, pp. 210--227, 2009.

\bibitem{Wagner12}
A.~Wagner, J.~Wright, A.~Ganesh, Z.~Zhou, H.~Mobahi, and Y.~Ma, ``Toward a
  practical face recognition system: Robust alignment and illumination by
  sparse representation,'' \emph{IEEE Transactions on Pattern Analysis and
  Machine Intelligence}, vol.~34, no.~2, pp. 372--386, 2012.

\bibitem{Yang10}
A.~Y. Yang, A.~Ganesh, S.~S. Sastry, and Y.~Ma, ``Fast l1-minimization
  algorithms and an application in robust face recognition: A review,'' in
  \emph{ICIP}, 2010, pp. 1849--1852.

\bibitem{Yang11a}
M.~Yang, L.~Zhang, J.~Yang, and D.~Zhang, ``Robust sparse coding for face
  recognition,'' in \emph{CVPR}, 2011.

\bibitem{Ma15PR}
J.~Ma, J.~Zhao, Y.~Ma, and J.~Tian, ``Non-rigid visible and infrared face
  registration via regularized gaussian fields criterion,'' \emph{Pattern
  Recognit.}, vol.~48, no.~3, pp. 772--784, 2015.

\bibitem{Wright10}
J.~Wright, Y.~Ma, J.~Mairal, G.~Sapiro, T.~S. Huang, and S.~Yan, ``Sparse
  representation for computer vision and pattern recognition,''
  \emph{Proceedings of the IEEE}, vol.~98, no.~6, pp. 1031--1044, 2010.

\bibitem{tan2006face}
X.~Tan, S.~Chen, Z.-H. Zhou, and F.~Zhang, ``Face recognition from a single
  image per person: A survey,'' \emph{Pattern recognition}, vol.~39, no.~9, pp.
  1725--1745, 2006.

\bibitem{Shi11}
Q.~Shi, A.~Eriksson, A.~van~den Hengel, and C.~Shen, ``Is face recognition
  really a compressive sensing problem?'' in \emph{CVPR}, 2011.

\bibitem{Zhang11}
L.~Zhang, M.~Yang, and X.~Feng, ``Sparse representation or collaborative
  representation: Which helps face recognition?'' in \emph{ICCV}, 2011.

\bibitem{Zhu10}
P.~Zhu, L.~Zhang, Q.~Hu, and S.~C. Shiu, ``Multi-scale patch based
  collaborative representation for face recognition with margin distribution
  optimization,'' in \emph{ECCV}, 2012.

\bibitem{Jiang14face}
J.~Jiang, R.~Hu, C.~Liang, Z.~Han, and C.~Zhang, ``Face image super-resolution
  through locality-induced support regression,'' \emph{Signal Process.}, vol.
  103, pp. 168--183, 2014.

\bibitem{Jiang17srlsp}
J.~Jiang, C.~Chen, J.~Ma, Z.~Wang, Z.~Wang, and R.~Hu, ``Srlsp: A face image
  super-resolution algorithm using smooth regression with local structure
  prior,'' \emph{IEEE Trans. Multimedia}, vol.~19, no.~1, pp. 27--40, 2017.

\bibitem{Deng12}
W.~Deng, J.~Hu, and J.~Guo, ``Extended {SRC}: Undersampled face recognition via
  intraclass variant dictionary,'' \emph{IEEE Transactions on Pattern Analysis
  and Machine Intelligence}, vol.~34, no.~9, pp. 1864--1870, 2012.

\bibitem{Yang13}
M.~Yang, L.~V. Gool, and L.~Zhang, ``Sparse variation dictionary learning for
  face recognition with a single training sample per person,'' in \emph{ICCV},
  2013.

\bibitem{Zhuang13}
L.~Zhuang, A.~Y. Yang, Z.~Zhou, S.~S. Sastry, and Y.~Ma, ``Single-sample face
  recognition with image corruption and misalignment via sparse illumination
  transfer,'' in \emph{CVPR}, 2013, pp. 3546--3553.

\bibitem{Zhuang14}
L.~Zhuang, T.-H. Chan, A.~Y. Yang, S.~S. Sastry, and Y.~Ma, ``Sparse
  illumination learning and transfer for single-sample face recognition with
  image corruption and misalignment,'' \emph{International Journal of Computer
  Vision}, 2014.

\bibitem{Wei15}
C.-P. Wei and Y.-C.~F. Wang, ``Undersampled face recognition via robust
  auxiliary dictionary learning,'' \emph{IEEE Transactions on Image
  Processing}, vol.~24, no.~6, pp. 1722--1734, 2015.

\bibitem{Deng13}
W.~Deng, J.~Hu, and J.~Guo, ``In defense of sparsity based face recognition,''
  in \emph{CVPR}, 2013, pp. 399--406.

\bibitem{Yang11}
M.~Yang, L.~Zhang, X.~Feng, and D.~Zhang, ``Fisher discrimination dictionary
  learning for sparse representation,'' in \emph{ICCV}, 2011.

\bibitem{Yang14}
------, ``Sparse representation based fisher discrimination dictionary learning
  for image classification,'' \emph{International Journal of Computer Vision},
  vol. 109, no.~3, pp. 209--232, 2014.

\bibitem{Ma12}
L.~Ma, C.~Wang, B.~Xiao, and W.~Zhou, ``Sparse representation for face
  recognition based on discriminative low-rank dictionary learning,'' in
  \emph{CVPR}, 2012.

\bibitem{Li13}
L.~Li, S.~Li, and Y.~Fu, ``Discriminative dictionary learning with low-rank
  regularization for face recognition,'' in \emph{Automatic Face and Gesture
  Recognition (FG)}, 2013.

\bibitem{Li14}
------, ``Learning low-rank and discriminative dictionary for image
  classification,'' \emph{Image and Vision Computing}, vol.~32, no.~10, pp.
  814--823, 2014.

\bibitem{Zhang13a}
Y.~Zhang, Z.~Jiang, and L.~S. Davis, ``Learning structured low-rank
  representations for image classification,'' in \emph{CVPR}, 2013.

\bibitem{Chi12}
Y.~Chi and F.~Porikli, ``Connecting the dots in multi-class classification:
  From nearest subspace to collaborative representation,'' in \emph{CVPR},
  2012, pp. 3602--3609.

\bibitem{Gao14}
S.~Gao, K.~Jia, L.~Zhuang, and Y.~Ma, ``Neither global nor local: Regularized
  patch-based representation for single sample per person face recognition,''
  \emph{International Journal of Computer Vision}, 2014.

\bibitem{Yan09}
S.~Yan and H.~Wang, ``Semi-supervised learning by sparse representation,'' in
  \emph{SDM}, 2009, pp. 792--801.

\bibitem{He11}
R.~He, W.-S. Zheng, B.-G. Hu, and X.-W. Kong, ``Nonnegative sparse coding for
  discriminative semi-supervised learning,'' in \emph{CVPR}, 2011, pp.
  2849--2856.

\bibitem{Zhuang12}
L.~Zhuang, H.~Gao, Z.~Lin, Y.~Ma, X.~Zhang, and N.~Yu, ``Non-negative low rank
  and sparse graph for semi-supervised learning,'' in \emph{CVPR}, 2012.

\bibitem{Zhuang15}
L.~Zhuang, S.~Gao, J.~Tang, J.~Wang, Z.~Lin, Y.~Ma, and N.~Yu, ``Constructing a
  non-negative low-rank and sparse graph with data-adaptive features,''
  \emph{IEEE Transactions on Image Processing}, 2015.

\bibitem{Wang12}
X.~Wang and X.~Tang, ``Bayesian face recognition based on gaussian mixture
  models,'' in \emph{ICPR}, 2004.

\bibitem{Zhu09}
X.~Zhu and A.~B. Goldberg, \emph{Introduction to Semi-Supervised
  Learning}.\hskip 1em plus 0.5em minus 0.4em\relax Morgan \& Claypool, 2009,
  ch.~3, pp. 21--35.

\bibitem{ma2015robust}
J.~Ma, H.~Zhou, J.~Zhao, Y.~Gao, J.~Jiang, and J.~Tian, ``Robust feature
  matching for remote sensing image registration via locally linear
  transforming,'' \emph{IEEE Transactions on Geoscience and Remote Sensing},
  vol.~53, no.~12, pp. 6469--6481, 2015.

\bibitem{Gao14_nonrigid}
Y.~Gao, J.~Ma, J.~Zhao, J.~Tian, and D.~Zhang, ``A robust and outlier-adaptive
  method for non-rigid point registration,'' \emph{Pattern Analysis and
  Applications}, vol.~17, no.~2, pp. 379--388, 2014.

\bibitem{Ma13PR}
J.~Ma, J.~Zhao, J.~Tian, X.~Bai, and Z.~Tu, ``Regularized vector field learning
  with sparse approximation for mismatch removal,'' \emph{Pattern Recognit.},
  vol.~46, no.~12, pp. 3519--3532, 2013.

\bibitem{Ma16}
J.~Ma, J.~Zhao, and A.~L. Yuille, ``Non-rigid point set registration by
  preserving global and local structures,'' \emph{IEEE Transactions on Image
  Processing}, 2016.

\bibitem{Kan13}
M.~Kan, S.~Shan, Y.~Su, D.~Xu, and X.~Chen, ``Adaptive discriminant learning
  for face recognition,'' \emph{Pattern Recognition}, vol.~46, no.~9, pp.
  2497--2509, 2013.

\bibitem{Su10}
Y.~Su, S.~Shan, X.~Chen, and W.~Gao, ``Adaptive generic learning for face
  recognition from a single sample per person,'' in \emph{CVPR}, 2010.

\bibitem{Martinez98}
A.~Martinez and R.~Benavente, ``The {AR} face database,'' CVC, Tech. Rep.~24,
  1998.

\bibitem{Gross09}
R.~Gross, I.~Matthews, J.~F. Cohn, T.~Kanade, and S.~Baker, ``Multi-{PIE},''
  \emph{Image and Vision Computing}, vol.~28, no.~5, pp. 807--813, 2010.

\bibitem{Gao08}
W.~Gao, B.~Cao, S.~Shan, X.~Chen, D.~Zhou, X.~Zhang, and D.~Zhao, ``The
  cas-peal large-scale chinese face database and baseline evaluations,''
  \emph{IEEE Transactions on Systems, Man and Cybernetics, Part A: Systems and
  Humans}, vol.~38, no.~1, pp. 149--161, 2008.

\bibitem{LFWTech}
G.~B. Huang, M.~Ramesh, T.~Berg, and E.~Learned-Miller, ``Labeled faces in the
  wild: A database for studying face recognition in unconstrained
  environments,'' University of Massachusetts, Amherst, Tech. Rep. 07-49,
  October 2007.

\bibitem{Asif10}
M.~S. Asif and J.~Romberg, ``Dynamic updating for l1 minimization,'' \emph{IEEE
  Journal of selected topics in signal processing}, 2010.

\bibitem{Donoho08}
D.~Donoho and Y.~Tsaig, ``Fast solution of l1-norm minimization problems when
  the solution may be sparse,'' \emph{IEEE Transactions on Information Theory},
  vol.~54, no.~11, pp. 4789--4812, 2008.

\bibitem{Osborne00}
M.~Osborne, B.~Presnell, and B.~Turlach, ``A new approach to variable selection
  in least squares problems,'' \emph{IMA journal of numerical analysis},
  vol.~20, no.~3, p. 389, 2000.

\bibitem{Martinez01}
A.~Martinez and A.~C. Kak, ``{PCA} versus {LDA},'' \emph{IEEE Transactions on
  Pattern Analysis and Machine Intelligence}, vol.~23, no.~2, pp. 228--233,
  2001.

\bibitem{Caiprobabilistic16}
S.~Cai, L.~Zhang, W.~Zuo, and X.~Feng, ``A probabilistic collaborative
  representation based approach for pattern classification,'' in \emph{CVPR},
  2012.

\bibitem{Bishop06}
C.~M. Bishop, \emph{Pattern Recognition and Machine Learning}, M.~Jordan,
  J.~Kleinberg, and B.~Sch$\ddot{o}$lkopf, Eds.\hskip 1em plus 0.5em minus
  0.4em\relax New York: Springer, 2006.

\bibitem{Chang11}
C.-C. Chang and C.-J. Lin., ``Libsvm : a library for support vector machines.''
  \emph{ACM Transactions on Intelligent Systems and Technology}, vol.~2,
  no.~27, pp. 1--27, 2011.

\bibitem{Gao16}
Y.~Gao and A.~L. Yuille, ``Symmetry non-rigid structure from motion for
  category-specific object structure estimation,'' in \emph{ECCV}, 2016.

\bibitem{Gao17}
------, ``Exploiting symmetry and/or {M}anhattan properties for 3{D} object
  structure estimation from single and multiple images,'' in \emph{CVPR}, 2017.

\bibitem{Parkhi15}
O.~M. Parkhi, A.~Vedaldi, and A.~Zisserman, ``Deep face recognition,'' in
  \emph{Proceedings of the British Machine Vision}, 2015.

\bibitem{sun2014deep}
Y.~Sun, Y.~Chen, X.~Wang, and X.~Tang, ``Deep learning face representation by
  joint identification-verification,'' in \emph{Advances in Neural Information
  Processing Systems}, 2014, pp. 1988--1996.

\bibitem{sun2015deeply}
Y.~Sun, X.~Wang, and X.~Tang, ``Deeply learned face representations are sparse,
  selective, and robust,'' in \emph{Proceedings of the IEEE Conference on
  Computer Vision and Pattern Recognition}, 2015, pp. 2892--2900.

\bibitem{chen2013blessing}
D.~Chen, X.~Cao, F.~Wen, and J.~Sun, ``Blessing of dimensionality:
  High-dimensional feature and its efficient compression for face
  verification,'' in \emph{Proceedings of the IEEE Conference on Computer
  Vision and Pattern Recognition}, 2013, pp. 3025--3032.

\bibitem{taigman2014deepface}
Y.~Taigman, M.~Yang, M.~Ranzato, and L.~Wolf, ``Deepface: Closing the gap to
  human-level performance in face verification,'' in \emph{Proceedings of the
  IEEE Conference on Computer Vision and Pattern Recognition}, 2014, pp.
  1701--1708.

\bibitem{Huang2012a}
G.~B. Huang, M.~Mattar, H.~Lee, and E.~Learned-Miller, ``Learning to align from
  scratch,'' in \emph{NIPS}, 2012.

\bibitem{Viola04}
P.~Viola and M.~J. Jones, ``Robust real-time face detection,''
  \emph{International Journal of Computer Vision}, vol.~57, no.~2, pp.
  137--154, 2004.

\bibitem{Yang12}
M.~Yang, L.~Zhang, and D.~Zhang, ``Efficient misalignment-robust representation
  for real-time face recognition,'' in \emph{ECCV}, 2012, pp. 850--863.

\bibitem{Deng14}
W.~Deng, J.~Hu, J.~Lu, and J.~Guo, ``Transform-invariant pca: A unified
  approach to fully automatic face alignment, representation, and
  recognition,'' \emph{IEEE Transactions on Pattern Analysis and Machine
  Intelligence}, vol.~36, no.~6, pp. 1275--1284, 2014.

\bibitem{ma2016infrared}
J.~Ma, C.~Chen, C.~Li, and J.~Huang, ``Infrared and visible image fusion via
  gradient transfer and total variation minimization,'' \emph{Information
  Fusion}, vol.~31, pp. 100--109, 2016.

\bibitem{ma2015robust2}
J.~Ma, W.~Qiu, J.~Zhao, Y.~Ma, A.~L. Yuille, and Z.~Tu, ``Robust l2e estimation
  of transformation for non-rigid registration.'' \emph{IEEE Trans. Signal
  Processing}, vol.~63, no.~5, pp. 1115--1129, 2015.

\end{thebibliography}

%







\vspace{-17 mm}
\begin{IEEEbiography}[{\includegraphics[width=1in,height=1.25in,clip,keepaspectratio]{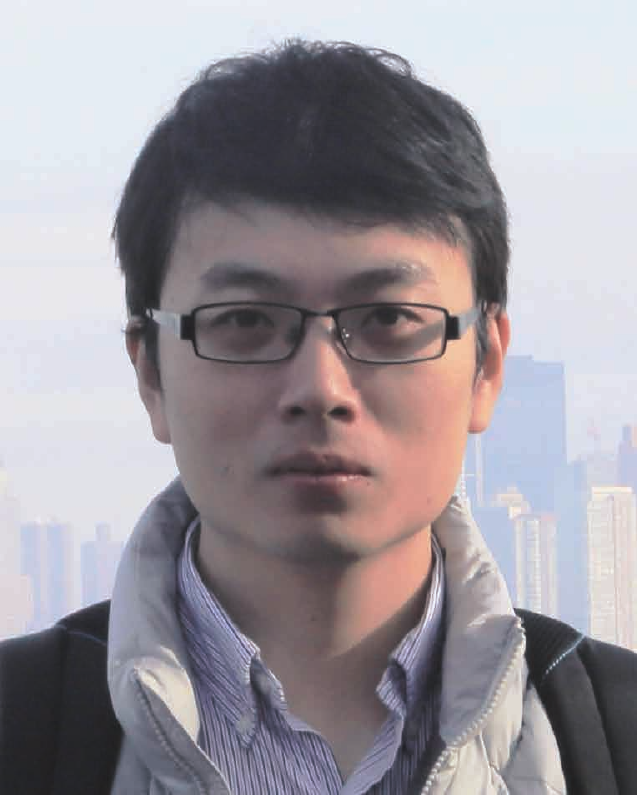}}]{Yuan Gao}
received the B.S. degree in biomedical engineering and the M.S. degree in pattern recognition and intelligent systems from the Huazhong University of Science and Technology, Wuhan, China, in 2009 and 2012, respectively. He completed his Ph.D. in Electronic Engineering, City University of Hong Kong, Kowloon, Hong Kong, in 2016. Currently, he is a computer vision researcher in Tencent AI Lab. He was a visiting student with the Department of Statistics, University of California, Los Angeles in 2015. His research interests include computer vision, pattern recognition, and machine learning.

\end{IEEEbiography}

\vspace{-15 mm}
\begin{IEEEbiography}[{\includegraphics[width=1in,height=1.25in,clip,keepaspectratio]{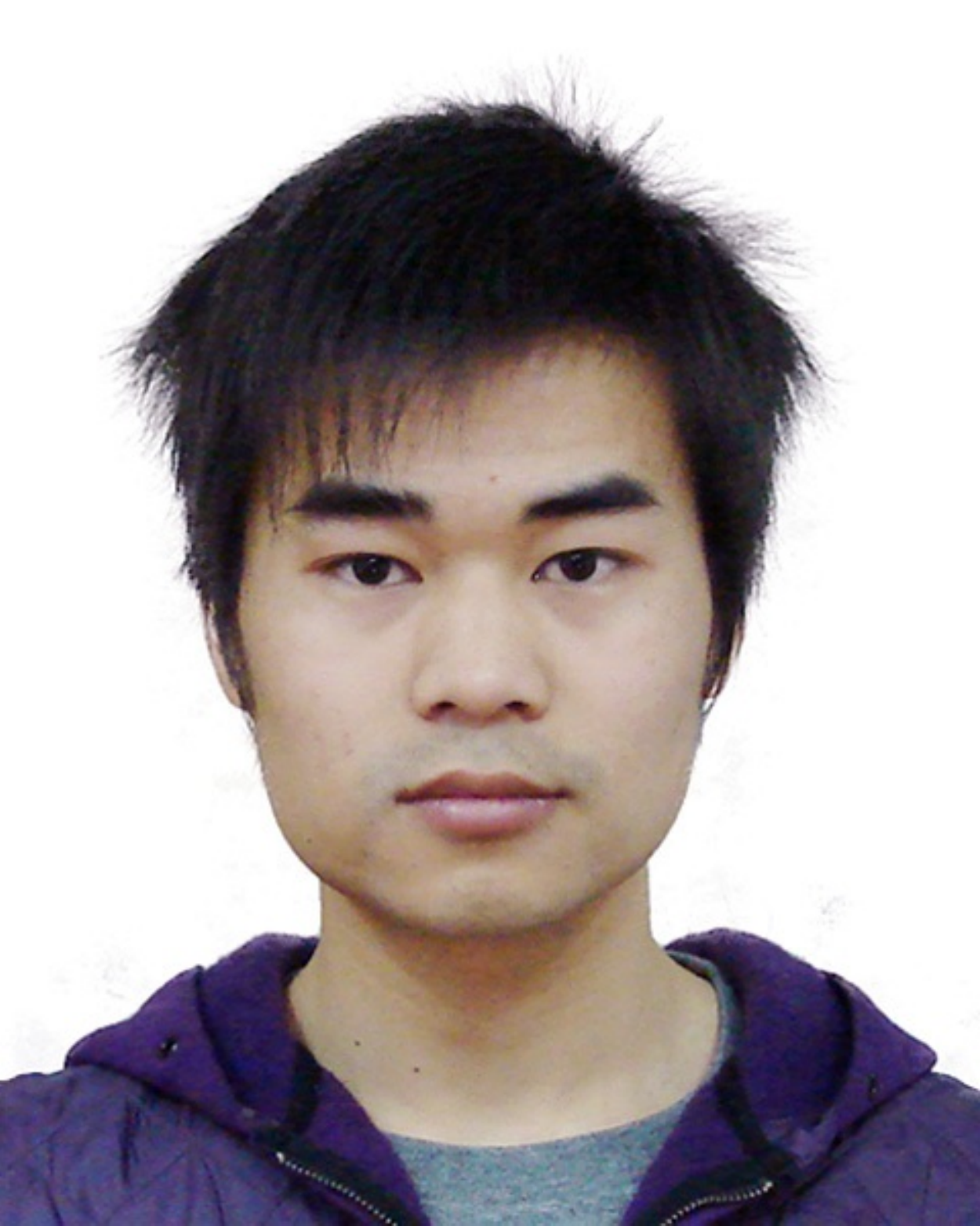}}]{Jiayi Ma}
received the B.S. degree from the Department of Mathematics, and the Ph.D. Degree from the School of Automation, Huazhong University of Science and Technology, Wuhan, China, in 2008 and 2014, respectively. From 2012 to 2013, he was an exchange student with the Department of Statistics, University of California, Los Angeles. He is now an Associate Professor with the Electronic Information School, Wuhan University, where he has been a Post-Doctoral during 2014 to 2015. His current research interests include in the areas of computer vision, machine learning, and pattern recognition.
\end{IEEEbiography}

\vspace{-15 mm}
\begin{IEEEbiography}[{\includegraphics[width=1in,height=1.25in,clip,keepaspectratio]{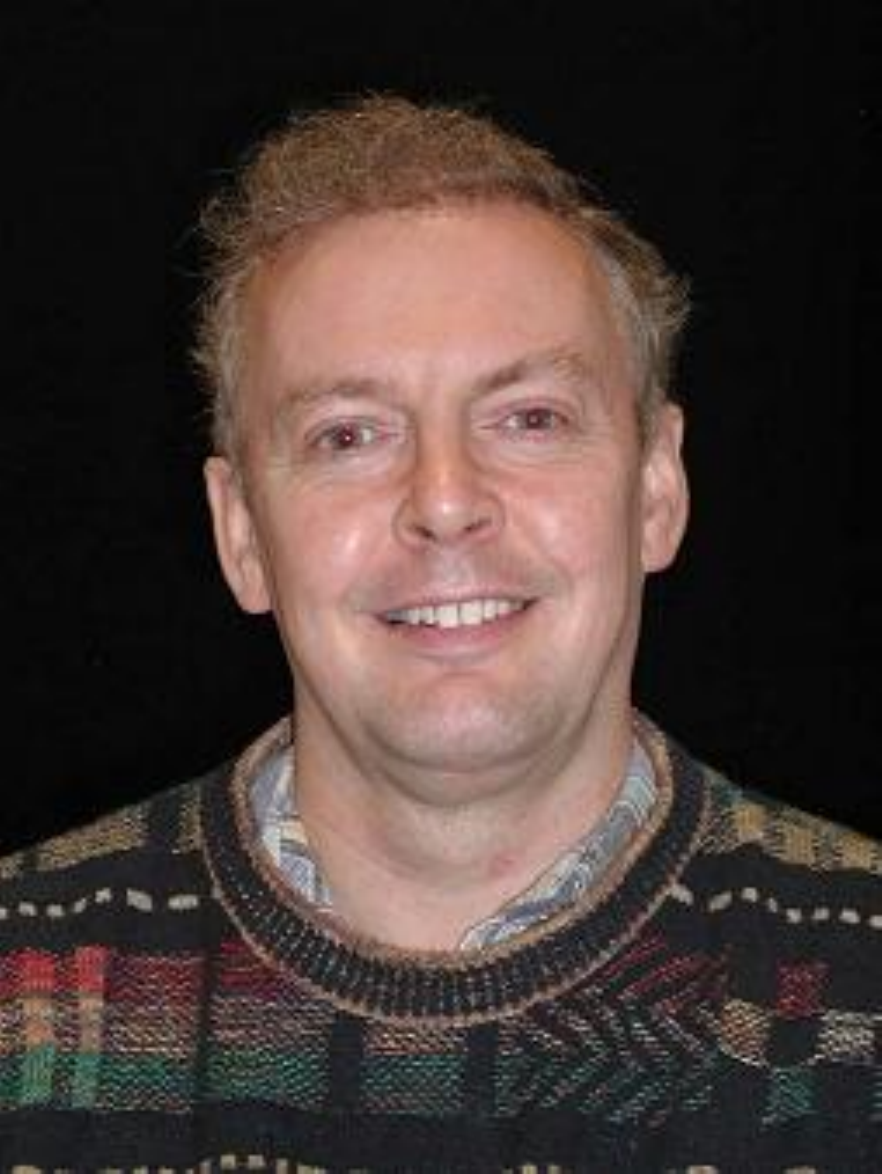}}]{Alan L. Yuille}
received his B.A. in mathematics from the University of Cambridge in 1976, and completed his Ph.D. in theoretical physics at Cambridge in 1980 studying under Stephen Hawking. Following this, he held a postdoc position with the Physics Department, University of Texas at Austin, and the Institute for Theoretical Physics, Santa Barbara. He then joined the Artificial Intelligence Laboratory at MIT (1982-1986), and followed this with a faculty position in the Division of Applied Sciences at Harvard (1986-1995), rising to the position of associate professor. From 1995-2002 Alan worked as a senior scientist at the Smith-Kettlewell Eye Research Institute in San Francisco. From 2002-2016 he was a full professor in the Department of Statistics at UCLA with joint appointments in Psychology, Computer Science, and Psychiatry. In 2016 he became a Bloomberg Distinguished Professor in Cognitive Science and Computer Science at Johns Hopkins University. He has over two hundred peer-reviewed publications in vision, neural networks, and physics, and has co-authored two books: Data Fusion for Sensory Information Processing Systems (with J. J. Clark) and Two- and Three-Dimensional Patterns of the Face (with P. W. Hallinan, G. G. Gordon, P. J. Giblin and D. B. Mumford); he also co-edited the book Active Vision (with A. Blake). He has won several academic prizes and is a Fellow of IEEE.
\end{IEEEbiography}

\end{document}